\documentclass[11pt,reqno]{amsart}
\usepackage[top=1.2in, bottom=1.2in, left=1.2in, right=1.2in]{geometry}
\usepackage{amsaddr}
\usepackage{amsfonts, amssymb, amsmath, dsfont, mathtools}%, enumerate, bm, xspace, graphicx, caption, subcaption}
\usepackage{amsthm}
\usepackage{mathrsfs} %Calligraphic fonts. Used in \CC here. 
\usepackage[linktoc=page, colorlinks, linkcolor=blue, citecolor=blue]{hyperref}
\usepackage{enumerate}
\usepackage{float}
\usepackage{graphicx}
\usepackage{xcolor}
\usepackage{subcaption} 
\usepackage[font=small]{caption} 
\numberwithin{equation}{section}

\newcommand{\be}{\begin{equation}}
\newcommand{\ee}{\end{equation}}

\def \N {\mathbb{N}}

\def \R {\mathbb{R}}

\newtheorem{theorem}{Theorem}[section]
\newtheorem{proposition}[theorem]{Proposition}

\newtheorem{definition}[theorem]{Definition}

\theoremstyle{remark}
\newtheorem{remark}[theorem]{Remark}

\begin{document}

\title{A Theory of Synaptic Neural Balance: From Local to Global Order}
%%should it be synaptic neural balance
%%\title{BiLU and Homogeneous Activation Functions and Generalized Neural Balance }
\author{Pierre Baldi, Antonios Alexos, Ian Domingo, Alireza Rahmansetayesh
}
\address{Department of Computer Science, University of California, Irvine}
% \email{pfbaldi@uci.edu, aalexos@uci.edu, idomingo@uci.edu, alireza.setayesh77@gmail.com} 

% Author 1
% \author{Pierre Baldi}
% \address{Department of Computer Science, University of California, Irvine}
% \email{pfbaldi@uci.edu}

\date{\today}

% \address{Department of Computer Science, University of California, Irvine}
% \email{pfbaldi@uci.edu}
% \address{Department of Computer Science, University of California, Irvine}
% \email{aalexos@uci.edu}
% \address{Department of Computer Science, University of California, Irvine}
% \email{idomingo@uci.edu}
% \address{Department of Computer Science, University of California, Irvine}
% \email{alireza.setayesh77@gmail.com}

\thanks{Work in part supported by ARO grant 76649-CS to P.B.}
%%, and U.S. Air Force grant FA9550-18-1-0031 to R. V}

\date{\today}

\begin{abstract}
We develop a general theory of synaptic neural balance and how it can emerge or be enforced in neural networks. For a given additive cost function $R$ (regularizer), a neuron is said to be in balance if the total cost of its input weights is equal to the total cost of its output weights. The basic example is provided by feedforward networks of ReLU units trained with $L_2$ regularizers, which exhibit balance after proper training. The theory explains this phenomenon and extends it in several directions. The first direction is the extension to bilinear and other activation functions. The second direction is the extension to more general regularizers, including all $L_p$ ($p>0$) regularizers. The third direction is the extension to non-layered architectures, recurrent architectures, convolutional architectures, as well as architectures with mixed activation functions. Gradient descent on the error function alone does not converge in general to a balanced state, where every neuron is in balance, even when starting from a balanced state. However, gradient descent on the regularized error function ought to converge to a balanced state, and thus network balance can be used to assess learning progress. The theory is based on two local neuronal operations: scaling which is commutative, and balancing which is not commutative. Finally, and most importantly, given any initial set of weights, when local balancing operations are applied to each neuron in a stochastic manner, global order always emerges through the convergence of the stochastic balancing algorithm to the same unique set of balanced weights. The reason for this convergence is the existence of an underlying strictly convex optimization problem where the relevant variables are constrained to a linear, only architecture-dependent, manifold. Simulations show that balancing neurons prior to learning, or during learning in alternation with gradient descent steps, can improve learning speed and final performance thereby expanding the arsenal of available training tools. Scaling and balancing operations are entirely local and thus physically plausible in biological and neuromorphic networks.
\end{abstract}

\maketitle

\setcounter{tocdepth}{1}
\tableofcontents

\section{Introduction}
%===============
 One of the most common complaints against neural networks is that they provide at best ``black-box'' solutions to problems.  In particular, when large neural networks are trained on complex tasks, they produce large arrays of synaptic weights that have no clear structure and are difficult to interpret. Thus finding any kind of structure in the weights of large neural networks is of great interest. Here we study a particular kind of structure we call synaptic neural balance
 and the conditions under which it emerges. 
Neural balance in general refers to some kind of homeostatic process that facilitates information processing in neural systems. Synaptic neural balance is different from other forms of neural balance which are based on the connectivity patterns or the activities of neurons, such as the biological notion of balance 
between excitation and inhibition 
\cite{froemke2015plasticity,field2020heterosynaptic,howes2022integrating,kim2022neurotransmitter,shirani2023physiological}.
In contrast, we use the term synpatic neural balance to  refer to any systematic relationship between the input and output synaptic weights of individual neurons, or layers of neurons. Here we consider the case where the cost of the input weights is equal to the cost of the output weights, where the cost is defined by some cost function or regularizer.  
 One of the most basic examples of such a relationship is when the sum of the squares of the input weights of a neuron is equal to the sum of the squares of its output weights.

\noindent
{\bf Basic Example:} The basic example where this happens is with a neuron with a ReLU activation function inside a network trained to minimize an error function with $L_2$ regularization. If we multiply the incoming weights of the neuron by some $\lambda>0$  and divide the outgoing weights of the neuron by the same $\lambda$, it is easy to see that this double scaling operation does not affect in any way the contribution of the neuron to the rest of the network. Thus, any component of the error function that depends only on the input-output function of the network is unchanged. However, the value of the $L_2$ regularizer changes with $\lambda$ and we can ask what is the value of $\lambda$ that minimizes the corresponding contribution given by:

\be
\sum_{i \in IN} (\lambda w_i)^2 + \sum_{i \in OUT} ( w_i/\lambda)^2=\lambda^2A+\frac{1}{\lambda^2}B
\label{eq:basic1}
\ee
where $IN$ and $OUT$ denote the set of incoming and outgoing weights respectively, $A=\sum_{i \in IN} w_i^2$, and $B=\sum_{i \in OUT}w_i^2$. The product of the two terms on the right-hand side of 
Equation \ref{eq:basic1} is equal to $AB$ and does not depend on $\lambda$. Since of all the rectangles with the same area the square has the shortest perimeter, the minimum is achieved when these two terms are equal, which yields:  $(\lambda^*)^4=B/A$ for the optimal $\lambda^*$. The corresponding new set of weights, $v_i=\lambda^* w_i$ for the input weights and 
$v_i=w_i/\lambda^*$ for the outgoing weights, must be balanced:
$\sum_{i \in IN}v_i^2=\sum_{i \in OUT} v_i^2$. This is because the optimal scaling factor for these rescaled weights must be  $\lambda^*=1$. 
Furthermore, if an entire network of ReLU neurons is properly trained using a standard error function with an $L_2$ regularizer, at the end of training one observes a remarkable phenomenon: for each ReLU neuron, the norm of the incoming synaptic weights is approximately equal to the norm of the outgoing synaptic weights, i.e. every neuron is balanced. Our  goal here is to gain a deeper understanding of this basic example and of how general the phenomenon of synaptic balance is.

There have been isolated previous studies of this kind of balance \cite{du2018algorithmic,stock2022synaptic} from different perspectives or under special conditions, restricted to particular neuronal models, architectures, or cost functions. Many of these studies propose to favor balance by adding an extra term to the loss function forcing the difference between input and output synaptic costs to be close to zero, or the ratio between input and output synaptic costs to be close to one
 \cite{yang2022better}.
%%propose to replace the $L_2$ regularization term in the loss with the sum of products of l2 (not squared) norms of the input and output weights associated each ReLU activation. They also prove the equivalence between $L_2$ regularization and the proposed term. 
The work in \cite{stock2022synaptic} proposes a new
biologically plausible learning rule
to increase robustness of neural dynamics
in non-linear recurrent networks resulting in a form of synaptic balance.
In \cite{du2018algorithmic}, the authors show that if a deep network is initialized in a balanced state with respect to the sum of squares metric, and if training progresses with an infinitesimal learning rate, then  the balance is preserved throughout training. 
Related results are also described in  \cite{arora2018convergence}. 
Other studies have also explored symmetry and balance effects on training neural networks. For example, \cite{neyshabur2015path} shows that training with stochastic gradient descent does not work well in highly unbalanced neural networks. As a result, these authors propose a rescaling-invariant solution analyzed in  \cite{neyshabur2015norm}. In the same vein, other authors have proposed that learning in neural networks can be accelerated with rescaling transformations \cite{zhao2022symmetry,armenta2023neural} without focusing on synaptic balancing. In \cite{saul2023weight} multiplicative rescaling factors, one at each hidden unit,
are used to balance the weights with a proof of convergence, but not of uniqueness.

Here we take a different, more unified, approach aimed at uncovering the generality of synaptic neuronal balance phenomena, the learning conditions under which they occur, as well as new local balancing algorithms and their convergence properties.
We explain and study synaptic neural balance in its generality in terms of activation functions, regularizers, network architectures, and training stages.
 In particular, we systematically answer questions such as: Why does balance occur? Does it occur only with ReLU neurons? Does it occur only with $L_2$ regularizers?  Does it occur only in certain architectures, e.g., fully connected feedforward architectures, as opposed to locally connected, convolutional, or recurrent architectures?
 Does it occur only at the end of training? In the process of answering these questions, we introduce local scaling and balancing operations for individual neurons or entire neural layers. Furthermore, we show that when these local operations are applied stochastically, a global balanced state always emerges, and this state is {\it unique} and depends only on the initial weights, but not on the order in which the neurons are balanced. In one rare case, the state can be solved analytically (Appendix). In addition, we 
provide experimental results showing that balancing neurons, before or during training, can improve learning convergence and performance in both feedforward and recurrent networks. 

\section{Homogeneous and BiLU Activation Functions}
In this section, we generalize the basic example of the introduction from the standpoint of the activation functions. In particular, we consider homogeneous activation functions (defined below). 
The importance of homogeneity has been previously identified in somewhat different contexts \cite{neyshabur2015data}.
%%The exact definition of homogeneous in this context is given below. 
Intuitively, homogeneity is a form of linearity with respect to weight scaling and thus it is useful to motivate the concept of homogeneous activation functions by looking at other notions of linearity for activation functions. This will also be useful for Section \ref{sec:beyondrelu} where even more general classes of activation functions are considered. 

\subsection{Additive Activation Functions}

\begin{definition}
 A neuronal activation function $f: \R  \to \R$ is additively linear if and only if 
 $f(x+y)=f(x)+(f(y)$ for any real numbers $x$ and $y$.
\end{definition}

\begin{proposition}
 The class of additively linear activation functions is exactly equal to the class of linear activation functions, i.e., activation functions of the form $f(x)=ax$.   
\end{proposition}

\begin{proof}
Obviously linear activation functions are additively linear. Conversely, if $f$ is additively linear, the following three properties are true:
\par\noindent
(1)  One must have: $f(nx)=nf(x)$ and $f(x/n)=f(x)/n$ for any $x \in \R$ and any $n \in \N$. As a result, $f(n/m)=nf(1)/m$ for any integers $n$ and $m$ ($m \not = 0$).
\par\noindent
(2) Furthermore, $f(0+0)=f(0)+f(0)$ which implies: $f(0)=0$.
\par\noindent
(3) And thus   $f(x-x) = f(x)+f(-x)=0$, which in turn implies that $f(-x)=-f(x)$.
\par\noindent
From these properties, it is easy to see that $f$ must be continuous, with  $f(x)=xf(1)$, and thus $f$ must be linear.
\end{proof}

\subsection{Multiplicative Activation Functions}

\begin{definition}
 A neuronal activation function $f: \R  \to \R$ is multiplicative if and only if 
 $f(xy)=f(x)(f(y)$ for any real numbers $x$ and $y$.
\end{definition}

\begin{proposition}
\label{prop:multiplicative}
 The class of continuous multiplicative activation functions is exactly equal to the class of functions comprising the functions: $f(x)=0$ for every $x$, $f(x)=1$ for every $x$, and all the even and odd functions satisfying 
 $f(x)=x^c$  for  $x \geq 0$, where $c$ is any constant in $\R$.
\end{proposition}

\begin{proof}
It is easy to check the functions described in the proposition are multiplicative. Conversely, assume $f$ is multiplicative. 
For both $x=0$ and $x=1$, we must have $f(x)=f(xx)=f(x)f(x)$ and thus $f(0)$ is either 0 or 1, and similarly for $f(1)$. If $f(1)=0$, then for any $x$ we must have $f(x)=0$ because: $f(x)=f(1x)=f(1)f(x)=0$. Likewise, if $f(0)=1$, then for any $x$ we must have $f(x)=1$ because: $1=f(0)=f(0x)=f(0)f(x)=f(x)$.
Thus, in the rest of the proof, we can assume that $f(0)=0$ and $f(1)=1$.
By induction, it is easy to see that for any $x \geq 0$ we must have: 
$f(x^n)=f(x)^n$ and $f(x^{1/n})=(f(x))^{1/n}$ for any integer (positive or negative). As a result, for any $x \in \R$ and any integers $n$ and $m$ we must have:  $f(x^{n/m})=f(x)^{n/m}$. By continuity this implies that for any $x \geq 0$ and any $r \in R$, we must have: $f(x^r)=f(x)^r$.
Now there is some constant $c$ such that: $f(e)=e^c$.
And thus, for any $x>0$, $f(x)=f(e^{\log x})=[f(e)]^{\log x}=e^{c\log x}=x^c$.
To address negative values of $x$, note that we must have $f[(-1)(-1=f(1)=1 f(-1)^2$. Thus, $f(-1)$ is either equal to 1 or to -1. Since for any $x>0$ we have $f(-x)=f(-1)f(x)$, we see that if $f(-1)=1$ the function must be even ($f(-x)=f(x)=x^c$), and if $f(-1)=-1$ the function must be odd ($f(-x)=-f(x)$).
\end{proof}

\par\noindent
We will return to multiplicative activation function in a later section.

\subsection{Linearly Scalable Activation Functions}
\begin{definition}
A neuronal activation function $f: \R  \to \R$ is linearly scalable if and only if 
$f(\lambda x)=\lambda f(x)$ for every $\lambda \in \R$.
\end{definition}
\begin{proposition}
 The class of linearly scalable activation functions is exactly equal to the class of linear activation functions, i.e., activation functions of the form $f(x)=ax$.   
\end{proposition}

\begin{proof}
Obviously, linear activation functions are linearly scalable.  For the converse, if $f$ is linearly multiplicative we must have $f(\lambda x)=\lambda f(x)=xf(\lambda)$ for any $x$ and any $\lambda$. By taking $\lambda=1$, we get $f(x)=f(1)x$ and thus $f$ is linear.
\end{proof}

Thus the concepts of linearly additive or linearly scalable activation function are of limited interest since both of them are equivalent to the concept of linear activation function. A more interesting class is obtained if we consider linearly scalable activation functions, where the scaling factor $\lambda$  is constrained to be positive  ($\lambda >0$), also called homogeneous functions.

\subsection{Homogeneous Activation Functions}

\begin{definition}{(Homogeneous)}
A neuronal activation function $f: \R  \to \R$ is homogeneous  if and only if: 
$f(\lambda x)=\lambda f(x)$ for every $\lambda \in \R$ with $\lambda >0$.
\end{definition}

\begin{remark}
Note that if $f$ is homogeneous, $f (\lambda 0)=\lambda f(0)=f(0)$ for any $\lambda>0$ and thus $f(0)=0$.
Thus it makes no difference in the definition of homogeneous if we set $\lambda \geq 0$ instead of $\lambda >0)$. 
\end{remark}

\begin{remark}
 Clearly, linear activation functions are homogeneous. However, there exists also homogeneous functions that are non-linear, such as ReLU or leaky ReLU activation functions. 
\end{remark}

We now provide a full characterization of the class of homogeneous activation functions. 
    
\subsection{ BiLU Activation Functions}

We first define a new class of activation functions, corresponding to bilinear units 
(BiLU), consisting of two half-lines meeting at the origin. This class contains all the linear functions, as well as the ReLU and leaky ReLU functions, and many other functions. 

\begin{definition}{(BiLU)}
A neuronal activation function $f: \R  \to \R$ is bilinear (BiLU)  if and only if 
$f(x)=ax$ when $x<0$, and 
$f(x)=bx$ when $x \geq 0$, for some fixed parameters $a$ and $b$ in $\R$.
 \end{definition}

These include linear units ($a=b$), ReLU units ($a=0,b=1$), leaky ReLU ($a=\epsilon;b=1$) units, and symmetric linear units ($a=-b$), all of which can also be viewed as special 
cases of piece-wise linear units \cite{tavakoli2021splash}, with a single hinge. One advantage of ReLU and more generally BiLU neurons, which is very important during backpropagation learning, is that their derivative is very simple and can only take one of two values ($a$ or $b$).

\begin{proposition}
A neuronal activation function $f: \R  \to \R$ is homogeneous if and only if it is a BiLU activation function. 
\end{proposition}

\begin{proof}
Every function in BiLU is clearly homogeneous. Conversely, 
any homogeneous function $f$ must satisfy:
(1) $f(0x)=0f(x)=f(0)=0$; (2)$f(x)= f(1x)=f(1)x$ for any positive $x$; and (3) $f(x)=f(-u)=f(-1)u=-f(-1)x$ for any negative $x$. Thus $f$ is in BiLU with $a=-f(-1)$ and $b=f(1)$.
\end{proof}

In Appendix A, we provide a simple proof that networks of BiLU neurons, even with a single hidden layer, have universal approximation properties.
In the next two sections, we introduce two fundamental neuronal operations, scaling and balancing, that can be applied to the incoming and outgoing synaptic weights of neurons with BiLU activation functions. 

\section{Scaling}

\begin{definition}{(Scaling)}
For any BiLU neuron $i$ in  network and any $\lambda >0$, we let 
$S_\lambda(i)$ denote the synaptic scaling operation by which the incoming connection  weights of neuron $i$ are multiplied by $\lambda$ and the outgoing connection weights of neuron $i$ are divided by $\lambda$.
\end{definition}

Note that because of the homogeneous property the scaling operation does not change how neuron $i$ affects the rest of the network. In particular, the input-output function of the overall network remains unchanged after scaling neuron $i$ bt any $\lambda >0$. 
Note also that scaling always preserves the sign of the synaptic weights to which it is applied, and 
the scaling operation can never convert a non-zero synaptic weight into a zero synaptic weight, or vice versa. 

As usual, the bias is treated here as an additional synaptic weight emanating from a unit clamped to the value one. Thus scaling is applied to the bias.

\begin{proposition} {(Commutativity of Scaling)}
Scaling operations applied to any pair of BiLU neurons $i$ and $j$  in a neural network commute:  $S_\lambda (i)S_{\mu}(j)=S_\mu(j)S_\lambda(i)$, in the sense that the resulting network weights are the same, regardless of the order in which the scaling operations are applied. Furthermore, for any BiLU neuron $i$:
$S_\lambda(i) S_\mu(i)=S_\mu(i) S_\lambda(i)   = S_{\lambda\mu}(i)$.
\end{proposition}
This is obvious. As a result, any set $I$ of BiLU neurons in a network can be scaled simultaneously or in any sequential order while leading to the same final configuration of synaptic weights. If we denote by $1,2, \dots,n$ the neurons in $I$, we can for instance  write:
$\prod_{i \in I}S_{\lambda_i}(i)=\prod_{\sigma(i) \in I}
S_{\lambda_{\sigma(i)}}(\sigma(i))$ for any permutation $\sigma$ of the neurons.  Likewise, we can collapse operations applied to the same neuron. For instance, we can write:
$S_5(1) S_2(2) S_3(1)  S_4(2)  =    S_{15}(1)S_8(2)=S_8(2)S_{15}(1)$

\begin{definition} {(Coordinated Scaling)}
   For any set $I$ of BiLU neurons in a network and any $\lambda >0$, we let $S_\lambda(I)$ denote the synaptic scaling operation by which all the neurons in $I$ are scaled by the same $\lambda$.
\end{definition}

\section{Balancing}

\begin{definition}{(Balancing)}
Given a BiLU neuron in a network, the balancing operation $B(i)$ is a particular scaling operation $B(i)=S_{\lambda^*}(i)$, where the scaling factor $\lambda^*$ is chosen to optimize a particular cost function, or regularizer, asociated with the incoming and outgoing weights of neuron $i$. 
\end{definition}    

 For now, we can imagine that this cost function is the usual $L_2$ (least squares) regularizer, but in the next section, we will consider more general classes of regularizers and study the corresponding optimization process.  For the $L_2$ regularizer, as shown in the next section, this optimization process results in a unique value of $\lambda^*$ such that sum of the squares of the incoming weights is equal to the sum of the squares of the outgoing weights, hence the term ``balance''. Note that obviously $B(B(i))=B(i)$ and that, as a special case of scaling operation, the balancing operation does not change how neuron $i$ contributes to the rest of the network, and thus it leaves the overall input-output function of the network unchanged. 

Unlike scaling operations, balancing operations in general do not commute as balancing operations (they still commute as scaling operations). Thus, in general, $B(i)B(j) \not = B(j)B(i)$.  This is because if neuron $i$ is connected to neuron $j$, balancing $i$ will change the connection between $i$ and $j$, and, in turn, this will change the value of the optimal scaling constant for neuron $j$ and vice versa.
However, if there are no non-zero connections between neuron $i$ and neuron $j$ then the balancing operations commute since each balancing operation will modify a different, non-overlapping, set of weights. 

\begin{definition}{(Disjoint neurons)}
Two neurons $i$ and $j$ in a neural network are said to be disjoint if there are no non-zero connections between $i$ and $j$. 
\end{definition}
Thus in this case $B(i)B(j) =S_{\lambda^*}(i)S_{\mu^*}(j)=
S_{\mu^*}(j) S_{\lambda^*}(i)=B(j)B(i)$. This can be extended to disjoint sets of neurons. 

\begin{definition} {(Disjoint Set of Neurons)}
   A set $I$ of neurons is said to be disjoint if for any pair $i$ and $j$ of neurons in $I $ there are no non-zero connections between  $i$ and $j$.
\end{definition}
For example, in a layered feedforward network, all the neurons in a layer form a disjoint set, as long as there are no intra-layer connections or, more precisely, no non-zero intra-layer connections. All the neurons in a disjoint set can be balanced in any order 
resulting in the same final set of synaptic weights. 
Thus we have:

\begin{proposition}
If we index by $1,2, \dots,n$ the neurons in a disjoint set $I$ of BiLU neurons in a network, we have:
$\prod_{i \in I}B(i)=\prod_{i \in I}S_{\lambda^*_i}(i)=\prod_{\sigma(i) \in I}
S_{\lambda^*_{\sigma(i)}}(\sigma(i))=\prod_{\sigma(i) \in I}B(\sigma(i))$ for any permutation $\sigma$ of the neurons.   
\end{proposition}

Finally, we can define the coordinated balancing of any set $I$ of BiLU neurons (disjoint or not disjoint).
\begin{definition}{(Coordinated Balancing)}
Given any set $I$ of BiLU neurons (disjoint or not disjoint) in a network, 
the coordinated balacing of these neurons, written as $B_{\lambda^*}(I)$,
corresponds to coordinated scaling  all the neurons in $I$ by the same factor $\lambda^*$, 
Where $\lambda^*$ minimizes the cost functions of all the weights, incoming and outgoing, associated with all the neurons in $I$.  
\end{definition}
\begin{remark}
While balancing corresponds to a full optimization of the scaling operation, it is also possible to carry a partial optimization of the scaling operation by choosing a scaling factor that reduces the corresponding contribution to the regularizer without minimizing it.    
\end{remark}

\section{General Framework and Single Neuron Balance}

%%\cite{du2018algorithmic}
%%\cite{parhi2022kinds} this was my strting point novak
%not cited here but should be in the future
%%\cite{neyshabur2015data}  discuss homogeneity
In this section, we generalize the kinds of regularizer to which the notion of neuronal synaptic balance can be applied, beyond the usual $L_2$ regularizer and derive the corresponding balance equations.  
Thus we consider a network (feedforward or recurrent) where the hidden units are BiLU units. The visible units can be partitioned into input units and output units. 
%%Take any hidden unit in the network, 
%%and let 
%%$v=v_1,\ldots,v_n$  denote its incoming weights, and $w=w_1,%%\ldots,w_m$  denote its outgoing weights.
For any hidden unit $i$, if we multiply all its incoming weights $IN(i)$ by some $\lambda>0$ and all its outgoing weights $OUT(i)$ by $1/\lambda$ the overall function computed by the network remains unchanged due to the BiLU homogeneity property. In particular, if there is an error function that depends uniquely on the input-output function being computed, 
this error remains unchanged by the introduction of the multiplier $\lambda$. However, if there is also a regularizer $R$ for the weights, its value is affected by $\lambda$ and one can ask what is the optimal value of $\lambda$ with respect to the regularizer, and what are the properties of the resulting weights. This approach can be applied to any regularizer. For most practical purposes, we can assume that the regularizer is continuous in the weights (hence in $\lambda$) and lower-bounded. Without any loss of generality, we can assume that it is lower-bounded by zero. If we want the minimum value to be achieved by some $\lambda>0$, we need to add some mild condition that prevents the minimal value to be approached as $\lambda \to 0^)$, or as $\lambda \to +\infty$.
For instance, it is enough if there is an interval $[a,b]$ with $0<a<b$
where $R$ achieves a minimal value $R_{min}$ and $R \geq R_{min}$ in the intervals $(0,a]$ and $[b,  +\infty)$. Additional (mild) conditions must be imposed if one wants the optimal value of $\lambda$ to be unique, or computable in closed form (see Theorems below). Finally, we want to be able to apply the balancing approach 

Thus, we consider overall regularized error functions, where the regularizer is very general, as long as it has an additive form with respect to the individual weights:
 
 \be
{\mathcal {E}}(W) = E(W) +R(W) \quad{\rm with} \quad R(W) = \sum_w g_w(w)
\label{eq:error100}
\ee
where  $W$ denotes all the weights in the network and $E(W)$ is typically the negative log-likelihood (LMS error in regression tasks, or cross-entropy error in classification tasks).
We assume that the $g_w$ are continuous, and lower-bounded by 0. To ensure the existence and uniqueness of minimum during the balancing of any neuron, We will assume that each function  $g_w$ depends only on the magnitude $\vert w \vert$ of the corresponding weight, and that 
$g_w$ is monotonically increasing from 0 to $+\infty$ ($g_w(0)=0$ and $\lim_{x \to + \infty} g_w(x)=+\infty$).
Clearly, 
$L_2, L_1$ and more generally all $L_p$ regularizers are special cases where, for $p>0$, $L^p$ regularization is defined by:
$R(W)= \sum _w  \vert w \vert^p$.

When indicated, we may require also that the functions $g_w$
be continuously differentiable, except perhaps at the origin in order to be able to differentiate the regularizer with respect to  the $\lambda$'s and derive closed form conditions for the corresponding optima. This is satisfied by all forms of $L_p$ regularization, for $p>0$. 

\begin{remark}
Often one introduces scalar multiplicative hyperparameters to balance the effect of $E$ and $R$, for instance in the form:  ${\mathcal{E}}=E + \beta R$. These cases are included in the framework above: multipliers like $\beta$ can easily be absorbed into the functions $g_w$ above.
\end{remark}

\begin{theorem}
\label{thm:generalbalance}
(General Balance Equation).
Consider a neural network with BiLU activation functions in all the hidden units and 
overall error function of the form:
\be
{\mathcal {E}} = E(W) +R(W) \quad{\rm with} \quad R(W) = \sum_w g_w(w)
\label{eq:error101}
\ee
where each function $g_w(w)$ is continuous, depends on the magnitude $\vert w \vert$ alone, and grows monotonically from 
$g_w(0)=0$ to $g_w(+\infty)=+\infty$. For any setting of the weights $W$ and any hidden unit $i$ in the network and any $\lambda>0$ we can multiply the incoming weights of $i$ by $\lambda$ and the outgoing weights of $i$ by $1/\lambda$ without changing the overall error $E$. Furthermore, there exists a unique value $\lambda^*$ where the corresponding weights $v$
($ v=\lambda^* w$ for incoming weights, $v=w/\lambda^*$ for the outgoing weights) achieve the balance equation:

\be
\sum_{v \in IN(i)}g_w(v)=\sum_{w \in OUT(i)} g_w(v)
\label{eq:balance100}
\ee
\end{theorem}

\begin{proof}
Under the assumptions of the theorem, $E$ is unchanged under the rescaling of the incoming and outgoing weights of unit $i$ due to the homogeneity property of BiLUs. Without any loss of generality, we assume that at the beginning:
$\sum_{w \in IN(i)}g_w(w)  < \sum_{w \in OUT(i)} g_w(w)$. As we increase $\lambda$ from 1 to $+\infty$, by assumptions on functions $g_w$, the term $\sum_{w \in IN(i)} g_w(\lambda w) $ increases continuously from its initial value to $+\infty$, whereas the term $\sum_{w \in OUT(i)}g_w)w/\lambda)$
decreases continuously from its initial value to $0$.
Thus, there is a unique value $\lambda^*$ where the balance is realized. If at the beginning $\sum_{w \in IN(i)}g_w(w)  > \sum_{w \in OUT(i)} g_w(w)$, then the same argument is applied by decreasing $\lambda$ from 1 to 0.
\end{proof}

\begin{remark}
For simplicity, here and in other sections, we state the results in terms of a network of BiLU units. However, the same principles can be applied to networks where only a subset of neurons are in the BiLU class, simply by applying scaling and balancing operations to only those neurons. Furthermore, not all BiLU neurons need to have the same BiLU activation functios. For instance, the results still hold for a mixed network containing both ReLU and linear units.  
\end{remark}

\begin{remark}
In the setting of Theorem \ref{thm:generalbalance}, the balance equations do not necessarily minimize the corresponding regularization term.
%%$\sum_{w \in IN(i)}g_w(w)+ \sum_{w \in OUT(i)}g_w(w)$. 
This is addressed in the next theorem. 
\end{remark}

\begin{remark}
Finally, zero weights ($w=0$) can be ignored entirely as they play no role in scaling or balancing. Furthermore, if all the incoming or outgoing weights of a hidden unit were to be zero, it could be removed entirely from the network  
\end{remark}

\begin{theorem} 
\label{th:balanceopt}
(Balance and Regularizer Minimization)
We now consider the same setting as in Theorem \ref{thm:generalbalance}, but in addition we assume that the functions $g_w$ are continuously differentiable, except perhaps at the origin. Then, for any neuron, there exists at least one optimal value $\lambda^{*}$ that minimizes $R(W)$. Any optimal value must be a solution of the consistency equation:

\be 
\lambda^2 \sum_{w \in IN (i)} wg'_w(\lambda w) = \sum_{w \in OUT(i)} wg'_w(w/\lambda)
\label{eq:gen30}
\ee
Once the weights are rebalanced accordingly, the new weights must satisfy the generalized balance equation:

\be
\sum_{w \in IN(i)}w g'(w) = \sum_{w \in OUT(i)}wg'(w)
\label{eq:balance101}
\ee
In particular, if $g_w(w)=\vert w \vert^p$ for all the incoming and outgoing weights of neuron $i$, then the optimal value $\lambda^{*}$ is unique and equal to:

\be
\lambda^{*} =
\Bigl (  \frac{\sum_{w \in OUT(i)} \vert w \vert ^{p}}  
{\sum_{w \in IN (i)} \vert  w \vert^{p}}
\Bigr )^{1/2p}=\Bigl ( \frac{\vert\vert OUT(i)\vert\vert_p}{\vert\vert IN(i) \vert\vert_p} \Bigr )^{1/2}
\label{eq:lambdaopt4}
\ee
The decrease $\Delta R \geq 0$ in the value of the $L_p$ regularizer  $R=\sum_w \vert w \vert^p$ is given by:

\be
\Delta R =\biggl (  \bigl (  \sum_{w \in IN(i)} \vert w \vert^p   \bigr )^{1/2} - \bigl (  \sum_{w \in OUT(i)} \vert w\vert^p \bigr )^{1/2} \biggr )^2
\label{eq:deltaR1}
\ee
After balancing neuron $i$, its new weights satisfy the generalized $L_p$ balance equation:

\be
\sum_{w \in IN (i)} \vert w \vert^p = \sum_{w \in OUT(i)} \vert w \vert ^p
\label{eq:lpbalance}
\ee
\end{theorem}

\begin{proof}
Due to the additivity of the regularizer, the only component of the regularizer that depends on $\lambda$ has the form:

\be
R(\lambda) =\sum_{w \in IN (i)} g_w(\lambda w) + \sum_{w \in OUT(i)} g_w(w/\lambda)
\label{eq:gen1}
\ee
Because of the properties of the functions $g_w$, $R_\lambda$ is continously differentiable and strictly bounded below by 0. So it must have a minimum, as a function of $\lambda$ where its derivative is zero.
Its derivative with respect to $\lambda$ has the form:

\be 
R'(\lambda)=\sum_{w \in IN (i)} wg'_w(\lambda w) + \sum_{w \in OUT(i)} (-w/\lambda^2)g'_w(w/\lambda)
\label{eq:gen2}
\ee
Setting the derivative to zero, gives:

\be 
\lambda^2 \sum_{w \in IN (i)} wg'_w(\lambda w) = \sum_{w \in OUT(i)} wg'_w(w/\lambda)
\label{eq:gen3}
\ee
Assuming that the left-hand side is non-zero, which is generally the case, the optimal value for $\lambda$ must satisfy:
%%\textcolor{blue} {condition for min?}

\be
\lambda = \Bigl (  \frac{\sum_{w \in OUT(i)} wg'_w(w/\lambda)}  
{\sum_{w \in IN (i)} wg'_w(\lambda w)}
\Bigr )^{1/2}
\label{eq:lambdaopt1}
\ee
If the regularizing function is the same for all the incoming and outgoing weights ($g_w=g$), then the optimal value $\lambda$ must satisfy:

\be
\lambda = \Bigl (  \frac{\sum_{w \in OUT(i)} wg'(w/\lambda)}  
{\sum_{w \in IN (i)} wg'(\lambda w)}
\Bigr )^{1/2}
\label{eq:lambdaopt2}
\ee
In particular, if $g(w)=\vert w \vert^p$ then
$g(w)$ is differentiable except possibly at 0 and $g'(w)= s(w) p \vert w \vert ^{p-1}$, where $s(w)$ denotes the sign of the weight $w$.
Substituting in Equation \ref{eq:lambdaopt2}, the optimal rescaling $\lambda$ must satisfy:

\be
\lambda^* =
\Bigl (  \frac{\sum_{w \in OUT(i)} w s(w)\vert w \vert ^{p-1}}  
{\sum_{w \in IN (i)} w \vert  ws(w)  \vert^{p-1}}
\Bigr )^{1/2p}=
\Bigl (  \frac{\sum_{w \in OUT(i)} \vert w \vert ^{p}}  
{\sum_{w \in IN (i)} \vert  w \vert^{p}}
\Bigr )^{1/2p}=\Bigl ( \frac{\vert\vert OUT(i)\vert\vert_p}{\vert\vert IN(i) \vert\vert_p} \Bigr )^{1/2}
\label{eq:lambdaopt41}
\ee
At the optimum, no further balancing is possible, and thus $\lambda^{*}=1$. Equation \ref{eq:gen3} yields immediately the generalized balance equation to be satisfied at the optimum:

\be
\sum_{w \in IN(i)}w g'(w) = \sum_{w \in OUT(i)}wg'(w)
\label{eq:gen10}
\ee
In the case of $L_P$ regularization, it is easy to check by applying Equation \ref{eq:gen10}, or by direct calculation that:

\be
\sum_{w \in IN(i)} \vert\lambda^* w \vert^p= 
\sum_{w \in OUT(i)} \vert w/\lambda^*\vert^p
\label{eq:pbalance100}
\ee
which is the generalized balance equation. Thus after balancing neuron, the weights of neuron $i$ satisfy the $L_p$ balance (Equation \ref{eq:lpbalance}).
The change in the value of the regularizer is given by:

\be
\Delta R = \sum_{w \in IN(i)} \vert w \vert^p + \sum_{w \in OUT(i)} \vert w\vert^p
- \sum_{w \in IN(i)} \vert \lambda^{*}w \vert^p
- \sum_{w \in OUT(i)}  \vert w/\lambda^{*}\vert^p
\label{eq:decrease1}
\ee
By substituting $\lambda^{*}$ by its explicit value given by Equation \ref{eq:lambdaopt41} and collecting terms gives Equation 
\ref{eq:deltaR1}.
\end{proof}

\begin{remark}
The monotonicity of the functions $g_w$ is not needed to prove the first part of Theorem \ref{th:balanceopt}. It is only needed to prove uniqueness of $\lambda^*$ in the $L_p$ cases.
\end{remark}

\begin{remark}
Note that the same approach applies to the case where there are multiple additive regularizers. For instance  with both $L^2 $ and $L^1$ regularization, in this case the function $f$ has the form: 
$g_w(w)=\alpha w^2+\beta \vert w\vert$. Generalized balance 
%in the form of Equation \ref{eq:lambdaopt2} 
still applies.
It also applies to the case where different regularizers are applied in different disconnected portions of the network.
\end{remark}

\begin{remark}
The balancing of a single BiLU neuron has little to do with the number of connections. It applies equally to fully connected neurons, or to sparsely connected neurons.
\end{remark} 

%-------------

\section{Scaling and Balancing Beyond BiLU Activation Functions}
\label{sec:beyondrelu}

So far we have generalized ReLU activation functions to BiLU activation functions in the context of scaling and balancing operations with positive scaling factors. While in the following sections we will continue to work with BiLU activation functions, in this section we show that the scaling and balancing operations can be extended even further to other activation functions. The section can be skipped if one prefers to progress towards the main results on stochastic balancing.  

 Given a neuron with activation function $f(x)$, during scaling instead of multiplying and dividing by $\lambda>0$, we could 
 multiply the incoming weights by a function $g(\lambda)$ and divide the outgoing weights by a function $h(\lambda)$, as long as the activation function $f$ satisfies:

\be
f(g(\lambda)x)=h(\lambda) f(x)
\label{eq:general10}
\ee
for every $x\in \R$
to ensure that the contribution of the neuron to the rest of the network remains unchanged.
Note that if the activation function $f$ satisfies Equation \ref{eq:general10}, so does the activation function $-f$.
In Equation \ref{eq:general10}, $\lambda$ does not have to be positive--we will simply assume that $\lambda$ belongs to some open (potentially infinite) interval $(a,b)$. Furthermore, the functions $g$ and $h$ cannot be zero for $\lambda \in (a,b)$ since they are used for scaling. It is reasonable to assume that the functions $g$ and $h$ are continuous, and thus they must have a constant sign as $\lambda$ varies over $(a,b)$.

Now, taking $x=0$ gives $f(0)=h(\lambda)f(0)$ for every $\lambda\in (a,b)$, and thus either $f(0)=0$ or $h(\lambda)=1$ for every $\lambda \in (a,b)$. The latter is not interesting and thus we can assume that the activation function $f$ satisfies $f(0)=0$. Taking $x=1$ gives 
$f(g(\lambda))=h(\lambda)f(1)$ for every $\lambda$ in $(a,b)$.
For simplicity, let us assume that $f(x)=1$. Then, we have:
$f(g(\lambda))=h(\lambda)$ for every $\lambda$. Substituting in Equation \ref{eq:general10} yields:

\be
f(g(\lambda)x)=f(g(\lambda)) f(x)
\label{eq:genehomo100}
\ee
for every $x \in \R$ and every $\lambda \in (a,b)$. This relation is essentially the same as the relation that defines multiplicative activation functions over the corresponding domain
(see Proposition \ref{prop:multiplicative}), and thus we can identify a key family of solutions using power functions. Note that we can define a new parameter $\mu=g(\lambda)$, where $\mu$ ranges also over some positive or negative interval $I$ over which: $f(\mu x)=f(\mu)f(x)$.

\subsection{Bi-Power Units (BiPU)}

Let us assume that $\lambda >0$, $g(\lambda)=\lambda$ and $h(\lambda)=\lambda^c$ for some $c \in \R$.
Then the activation function must satisfy the equation: 

\be
f(\lambda x)=\lambda^c f(x)
\label{eq:BiPU}
\ee
for any $x \in \R$ and any $\lambda >0$. Note that if $f(x)=x^c$ we get a multiplicative activation function. 
More generally, these functions are characterized by the following proposition. 

\begin{figure}[ht]
	\centering
		\includegraphics[keepaspectratio=true,width=1.0\textwidth]{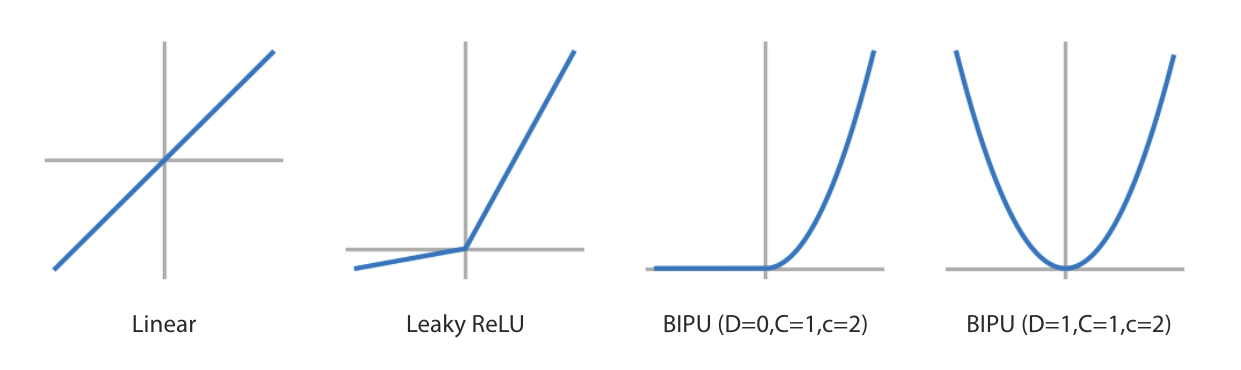}
		\caption{BiPU activation functions (Bi-Power-Units) as described in Equation \ref{eq:BiPU}}
		{
  }  
			\label{fig:activation_functions}
\end{figure}

\begin{proposition}
 The set of activation functions $f$ satisfying   $f(\lambda x)=\lambda^c f(x)$ for any $x\ \in \R$ and any $\lambda >0$ consist of the functions of the form:

\be
f(x)=
\begin{cases}
Cx^c \quad {\rm if} \quad x \geq 0 \\
Dx^c  \quad {\rm if} \quad x<0.
\end{cases}
\ee
where $c \in \R$, $C=f(1) \in R$, and $D=f(-1) \in \R$. We call these bi-power units (BiPU). If, in addition, we want $f$ to be continuous at $0$, we must have either $c>0$, or $c=0$ with $C=D$.
\end{proposition}
Given the general shape, these activation functions can be called BiPU (Bi-Power-Units). 
Note that in the general case where $c>0$, $C$ and $D$ do not need to be equal. In particular, one of them can be equal to zero, and the other one can be different from zero giving rise to ``rectified power units'' (Figure \ref{fig:activation_functions}).

\begin{proof}
 By taking $x=1$, we get $f(\lambda)=f(1)\lambda^c$ for any $\lambda>0$. Let $f(1)=C$. Then we see that for any $x>0$ we must have: $f(x)=Cx^c$. In addition, for every $\lambda >0$ we must have:
 $f(\lambda 0)=f(0)=\lambda^c f(0)$.  So if $c=0$, then $f(x)=C=f(1)$ for $x \geq 0$. If $c \not =0$, then $f(0)=0$. In this case, if we want the activation function to be continuous, then we see that we must have $c \geq 0$. So in summary for $x>0$  we must have $f(x)=f(1)x^c=Cx^c$. For the function to be right continuous at 0, we must have either $f(0)=f(1)=C$ with $c=0$ or $f(0)=0$ with $c>0$. We can now look at negative values of $x$. By the same reasoning, we have $f(\lambda (-1))=f(-\lambda)=\lambda^cf(-1)$ for any $\lambda>0$. Thus for any $x<0$ we must have: $f(x)=f(-1)\vert x \vert^c=D\vert x \vert^c$ where $D=f(-1)$. Thus, if $f$ is continuous, there are two possibilities. If $c=0$, then  we must have $C=f(1)=D(f-1) -$ and thus $f(x)=C$ everywhere. If $c \not =0$, then continuity requires that $c>0$. In this case $f(x)=Cx^c$ for $x \geq 0$ with $C=f(1)$, and $f(x)=Dx^c$ for $x<0$ with $f(-1)=D$. In all cases, it is easy to check directly that the resulting functions satisfy the functional equation 
 given by Equation \ref{eq:BiPU}.
\end{proof}

\subsection{Scaling BiPU Neurons}

A BiPU neuron can be scaled by multiplying its incoming weight by $\lambda>0$ and dividing its outgoing weights by $1/\lambda^c$. This will not change the role of the corresponding unit in the network, and thus it will not change the input-output function of the network. 

\subsection{Balancing BiPU Neurons}
As in the case of BiLU neurons, we balance a multiplicative neuron by asking what is the optimal scaling factor $\lambda$
that optimizes a particular regularizer. For simplicity, here we assume that the regularizer is in the $L_p$ class. Then we are interested in the value of $\lambda>0$ that minimizes the function:

\be
\lambda^p \sum_{w \in IN} \vert w \vert^p +
\frac{1}{\lambda^{pc}} \sum_{w \in OUT} \vert w \vert^p
\label{eq:balance1000}
\ee
A simple calculation shows that the optimal value of $\lambda$ is given by:

\be
\lambda^*=  \Bigl (   \frac {c \sum_{OUT}\vert w \vert^p}
{\sum_{IN} \vert w \vert^p} \Bigr )
^{1/p(c+1)}
\label{eq:balance1001}
\ee
Thus after balancing the weights, the neuron must satisfy the balance equation:

\be
 {c \sum_{OUT}\vert w \vert^p} = \sum_{IN} \vert w \vert^p
\label{eq:balance1002}
\ee
in the new weights $w$.

So far, we have focused on balancing individual neurons. In the next two sections, we look at balancing across all the units of a network. We first look at what happens to network balance when a network is trained by gradient descent and then at what happens to network balance when individual neurons are balanced iteratively in a regular or stochastic manner. 

\section{Network Balance: Gradient Descent}
\label{sec:SGD}
%%Balancing Neurons and Networks}

A natural question is whether gradient descent (or stochastic gradient descent) applied to a network of BiLU neurons, with or without a regularizer, converges to a balanced state of the network, where all the BiLU neurons are balanced. So we first consider the case where there is no regularizer (${\mathcal { E}}=E$). The results in \cite{du2018algorithmic}
may suggest that gradient descent may converge to a balanced state. In particular, they write that for any neuron $i$:

\be
\frac{d}{dt} \bigl ( \sum_{w \in IN(i)}  w^2  - \sum_{w \in OUT(i)} w^2
\bigr )=0
\ee
Thus the gradient flow exactly preserves the difference between the $L_2$ cost of the incoming and outgoing weights or, in other words, the derivative of the $L_2$ balance {\it deficit} is zero.
Thus if one were to start from a balanced state and use an infinitesimally small learning rate one ought to stay in a balanced state at all times. 

However, it must be noted that this result was derived for the $L_2$ metric only, and thus would not cover other $L_p$ forms of balance. Furthermore, it requires an infinitesimally small learning rate. In practice, when any standard learning rate is applied, we find that gradient descent does {\it not} converge to a balanced state (Figure 1). However, things are different when a regularizer term is included in the error functions as described in the following theorem. 

\begin{theorem}
Gradient descent in a network of BiLU units with error function ${\mathcal E}=E+R$ where $R$ has the properties described in Theorem \ref{th:balanceopt} (including all  $L_p$) must converge to a balanced state, where every BiLU neuron is balanced.
\end{theorem}

\begin{proof}
By contradiction, suppose that gradient descent converges to a state that is unbalanced and where the gradient with respect to all the weights is zero. Then there is at least one unbalanced neuron in the network. We can then multiply the incoming weights of such a neuron by $\lambda$ and the outgoing weights by $1/\lambda$ as in the previous section without changing the value of $E$. Since the neuron is not in balance, we can move $\lambda$ infinitesimally so as to reduce $R$, and hence $\mathcal E$. But this contradicts the fact that the gradient is zero.
\end{proof}

\begin{remark}
In practice, in the case of stochastic gradient descent applied to $E+R$, at the end of learning the algorithm may hover around a balanced state. If the state reached by the stochastic gradient descent procedure is not approximately balanced,  then learning ought to continue. In other words, the degree of balance could be used to monitor whether learning has converged or not. Balance is a necessary, but not sufficient, condition for being at the optimum. 
\end{remark}
\begin{remark}
If early stopping is being used to control overfitting, there is no reason for the stopping state to be balanced. However, the balancing algorithms described in the next section could be used to balance this state. 
\end{remark}

\section{Network Balance: Stochastic or Deterministic Balancing Algorithms}

In this section, we look at balancing algorithms where, starting from an initial weight configuration $W$, the BiLU neurons of a network are balanced iteratively according to some deterministic or stochastic schedule that periodically visits all the neurons.  We can also include algorithms where neurons are partitioned into groups (e.g. neuronal layers) and neurons in each group are balanced together. 
 
\subsection{Basic Stochastic Balancing}
The most interesting algorithm is when the BiLU neurons of a network are iteratively balanced in a purely stochastic manner. This algorithm is particularly attractive from the standpoint of physically implemented neural networks because the balancing algorithm is local and the updates occur randomly without the need for any kind of central coordination. 
%"orchestra conductor"
As we shall see in the following section, the random local operations remarkably lead to a unique form of global order. The proof for the stochastic case extends immediately to the deterministic case, where the BiLU neurons are updated in a deterministic fashion, for instance by repeatedly cycling through them according to some fixed order.  

\subsection{Subset Balancing (Independent or Tied)}
It is also possible to partition the BiLU neurons into non-overlapping subsets of neurons, and then balance each subset, especially when the neurons in each subset are disjoint of each other. In this case, one can balance all the neurons in a given subset, and repeat this subset-balancing operation subset-by-subset, again in a deterministic or stochastic manner. Because the  BiLU neurons in each subset are disjoint, it does not matter whether the neurons in a given subset are updated synchronously or sequentially (and in which order). Since the neurons are balanced independently of each other, this can be called independent subset balancing.
For example, in a layered feedforward network with no lateral connections,  each layer corresponds to a subset of disjoint neurons. The incoming and outgoing connections of each neuron are distinct from the incoming and outgoing connections of any other neuron in the layer, and thus the balancing operation of any neuron in the layer does not interfere with the balancing operation of any other neuron in the same layer. So this corresponds to independent layer balancing, 

As a side note, balancing a layer $h$, may disrupt the balance of layer $h+1$. However, balancing layer $h$ and $h+2$ (or any other layer further apart) can be done without interference of the balancing processes. This suggests also an alternating balancing scheme, where one alternatively balances all the odd-numbered layers, and all the evenly-numbered layers. 

Yet another variation is when the neurons in a disjoint subset are tied to each other in the sense that they must all share the same scaling factor $\lambda$. In this case, balancing the subset requires finding the optimal $\lambda$ for the entire subset, as opposed to finding the optimal $\lambda$ for each neuron in the subset. Since the neurons are balanced in a coordinated or tied fashion, this can be called coordinated or tied subset balancing. For example, tied layer balancing must use the same $\lambda$ for all the neurons in a given layer. 
It is easy to see that this approach leads to layer synaptic balance which has the form (for an $L_p$ regularizer):

\be
\sum_i \sum_{w \in IN (i)} \vert w \vert^p = \sum_i \sum_{w \in OUT(i)} \vert w \vert ^p
\label{eq:Coordinated_Layer_Balancing}
\ee
where $i$ runs over all the neurons in the layer. This does {\it not} necessarily imply that each neuron in the layer is individually balanced.
Thus neuronal balance for every neuron in a layer implies layer balance, but the converse is not true. Independent layer balancing will lead to layer balance. Coordinated layer balancing will lead to layer balance, but not necessarily to neuronal balance of each neuron in the layer. 
Layer-wise balancing, independent or tied,  can be applied to all the layers and in deterministic (e.g. sequential) or stochastic manner. 
Again the proof given in the next section for the basic stochastic algorithm can easily be applied to these cases (see also Appendix B). 

\subsection{Remarks about Weight Sharing and Convolutional Neural Networks}

Suppose that two connections share the same weight so that we must have: $ w_{ij}=w_{kl}$
at all times. In general, when the balancing algorithm is applied to neuron $i$ or $j$,
the weight $w_{ij}$ will change and the same change must be applied to $w_{kl}$. The latter may disrupt the balance of neuron $k$ or $l$. Furthermore, this may not lead to a decrease in the overall value of the regularizer $R$.
%%Thus in general a balancing algorithm applied to a network with shared weights may not converge.

The case of convolutional networks is somewhat special, since {\it all} the incoming weights of the neurons sharing the same convolutional kernel are shared. However, in general, the outgoing weights are not shared. Furthermore, certain operations like max-pooling are not homogeneous. So if one trains a CNN with $E$ alone, or even with $E+R$, one should not expect any kind of balance to emerge in the convolution units. However, all the other BiLU units in the network should become balanced by the same argument used for gradient descent above. The balancing algorithm applied to individual neurons, or the independent layer balancing algorithm, will not balance individual neurons sharing the same convolution kernel. The only balancing algorithm that could lead to some convolution layer balance, but not to individual neuronal balance, is the coordinated layer balancing, where the same $\lambda$ is used for all the neurons in the same convolution layer, provided that their activation functions are BiLU functions. 

We can now study the convergence properties of balancing algorithms.

\section{Convergence of Balancing Algorithms}

We now consider the basic stochastic balancing algorithm, where BiLU neurons are iteratively and stochastically balanced. It is essential to note that balancing a neuron $j$ may break the balance of another neuron $i$ to which $j$ is connected. Thus convergence of iterated balancing is not obvious.  
There are three key questions to be addressed for the basic stochastic algorithm, as well as all the other balancing variations. First, does the value of the regularizer converges to a finite value? Second, do the weights themselves converge to fixed finite values representing a balanced state for the entire network? And third, if the weights converge, do they always converge to the same values, irrespective of the order in which the units are being balanced? In other words, given an initial state $W$ for the network, is there a unique corresponding balanced state, with the same input-output functionalities?

\subsection{Notation and Key Questions}
For simplicity, we use a continuous time notation. 
After a certain time $t$ each neuron has been balanced a certain number of times. While the balancing operations are not commutative as balancing operations, they are commutative as scaling operations. Thus we can reorder the scaling operations and group them neuron by neuron so that, for instance, neuron $i$ has been scaled by the sequence of scaling operations:

\be
S_{\lambda_1^*}(i)S_{\lambda_2^*}(i)\ldots S_{\lambda_{n_{it}}^*}(i) = S_{\Lambda_i(t)}(i)
\label{eq:commute100}
\ee
where $n_{it}$ corresponds to the count of the last update of neuron $i$ prior to time $t$, and:

\be
\Lambda_i(t) = \prod_{1 \leq n \leq n_{it}} \lambda_n^*(i)
\label{eq:commute200}
\ee
For the input and output units, we can consider that their balancing coefficients $\lambda^*$ are always equal to 1 (at all times) and therefore
$\Lambda_i(t)=1$ for any visible unit $i$.

Thus, we first want to know if $R$ converges. Second, we want to know if the weights converge. This question can be split into two sub-questions: (1) Do the balancing factors $\lambda^*_n(i)$ converge to a limit as time goes to infinity. Even if the  $\lambda^*_n(i)$'s converge to a limit, this does not imply that the weights of the network converge to a limit. After a time $t$, the weight 
$w_{ij}(t)$ between neuron $j$ and neuron $i$ has the value
$w_{ij}\Lambda_i(t)/\Lambda_j(t)$, where $w_{ij}=w_{ij}(0)$ is the value of the weight at the start of the stochastic balancing algorithm. Thus: (2) Do the quantities 
$\Lambda_i(t)$ converge to finite values, different from 0?
And third, if the weights converge to finite values different from 0, are these values unique or not, i.e. do they depend on the details of the stochastic updates or not? These questions are answered by the following main theorem..

\subsection{Convergence of the Basic Stochastic Balancing Algorithm to a Unique Optimum}

\begin{theorem}
\label{thm:uniqueness}
(Convergence of Stochastic Balancing)
Consider a network of BiLU neurons with an error function 
${\mathcal E}(W)=E(W)+R(W)$ where $R$ satisfies the conditions of Theorem \ref{thm:generalbalance} including all  $L_p$ ($p>0$). Let $W$ denote the initial weights. When the neuronal stochastic balancing algorithm is applied throughout the network so that every neuron is visited from time to time, then $E(W)$ remains unchanged but $R(W)$ must converge to some finite value that is less or equal to the initial value, strictly less if the initial weights are not balanced. In addition, 
for every neuron $i$, $\lambda^*_i(t) \to 1$ and $\Lambda_i(t) \to \Lambda_i$ as $t \to \infty$, where $\Lambda_i $ is finite and $\Lambda_i > 0$ for every $i$. As a result, the weights themselves must converge to a limit $W'$ which is globally balanced, with $E(W)=E(W')$ and $R(W) \geq R(W')$, and with equality if only if $W$ is already balanced.
Finally, $W'$ is unique as it corresponds to the solution of a strictly convex optimization problem in the variables $L_{ij}= \log (\Lambda_i/\Lambda_j) $  with linear constraints of the form $\sum_\pi L_{ij}=0$ along any path $\pi$ joining an input unit to an output unit and along any directed cycle (for recurrent networks). Stochastic balancing projects to stochastic trajectories in the linear manifold that run from the origin to the unique optimal configuration. 
\end{theorem}

\begin{proof}
Each individual balancing operation leaves $E(W)$ unchanged because the BiLU neurons are homogeneous. Furthermore, each balancing operation reduces the regularization error $R(W)$, or leaves it unchanged. 
Since the regularizer is lower-bounded by zero, the value of the regularizer must approach a limit as the stochastic updates are being applied. 

For the second question, when neuron $i$ is balanced at some step, we know that the regularizer $R$ decreases by:

\be
\Delta R =\biggl (  \bigl (  \sum_{w \in IN(i)} \vert w \vert^p   \bigr )^{1/2} - \bigl (  \sum_{w \in OUT(i)} \vert w\vert^p \bigr )^{1/2} \biggr )^2
\label{eq:deltaR2}
\ee
If the convergence were to occur in a finite number of steps, then the coefficients $\lambda^*_i(t)$ must become equal and constant to 1 and the result is obvious. So we can focus on the case where the convergence does not occur in a finite number of steps (indeed this is the main scenario, as we shall see at the end of the proof). Since $\Delta R \to 0$, we must have:

\be 
\sum_{w\in IN(i)} \vert w \vert^p  \to 
\sum_{w \in OUT(i) }\vert w \vert^p
\label{eq:balance1010}
\ee
But from the expression for $\lambda^*$ (Equation \ref{eq:lambdaopt41}),
this implies that for every $i$, $\lambda^*_n(i) \to 1$ as time increases ($n \to \infty$). 
This alone is not sufficient to prove that $\Lambda_i(t)$ converges for every $i$ as $t \to \infty$. However, it is easy to see that $\Lambda_i(t)$ cannot contain a sub-sequence that approaches 0 or $\infty$ (Figure \ref{fig:Limits}). Furthermore, not only $\Delta R$ converges to 0, but the series $\sum \Delta R $ is convergent. This shows that, for every $i$,  $\Delta_i(t)$ must converge to a finite, non-zero value $\Delta_i$. Therefore all the weights must converge to fixed values given by $w_{ij} (0)\Lambda_i/\Lambda_j$.

\begin{figure}[ht]
	\centering
		\includegraphics[keepaspectratio=true,width=1.0\textwidth]{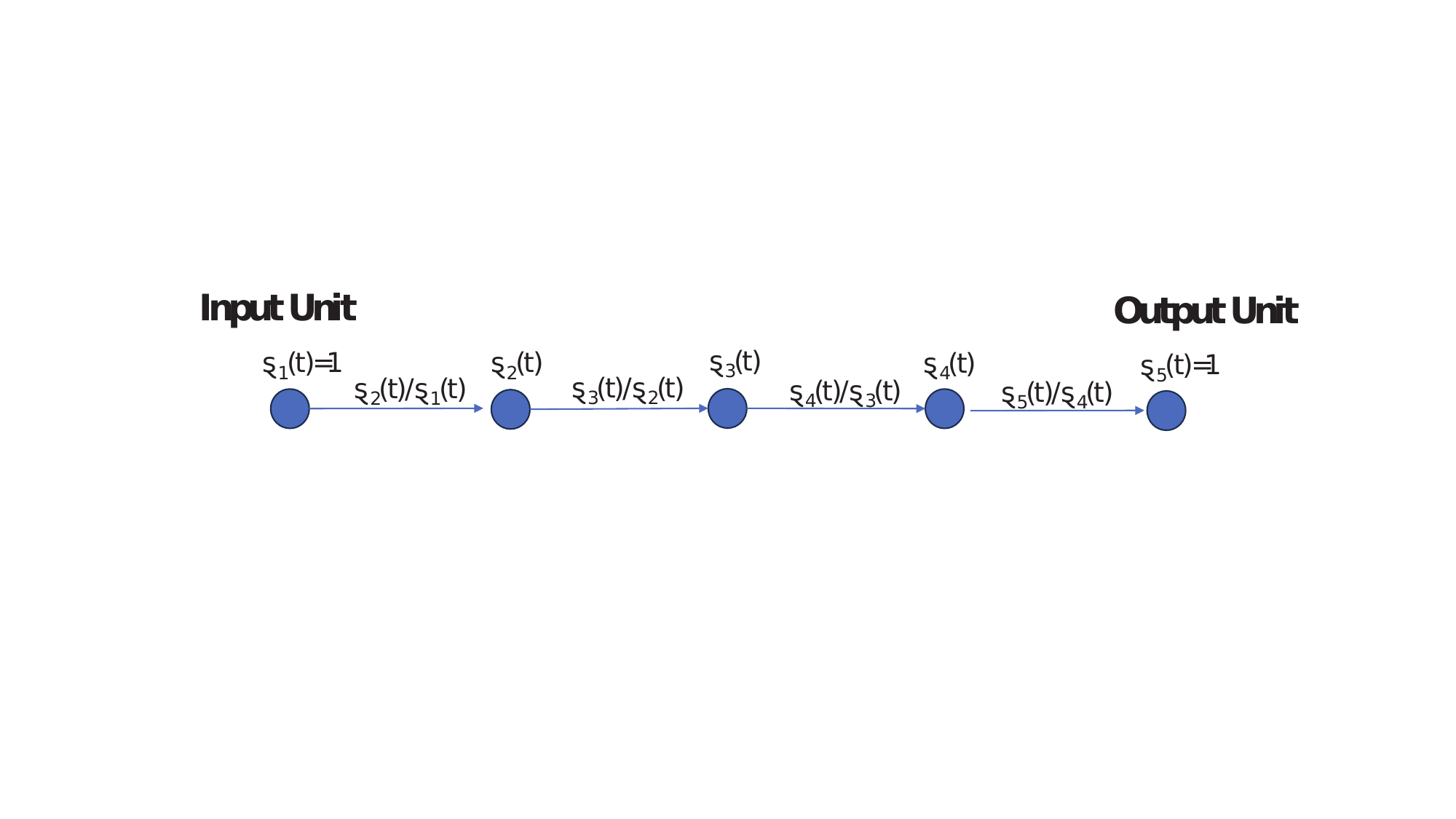}
  \vspace{-4cm}
		\caption[temp]
		{A path with three hidden BiLU units connecting one input unit to one output unit. During the application of the stochastic balancing algorithm, at time $t$ each unit $i$ has a cumulative scaling factor $\Lambda_i(t)$, and each directed edge from unit $j$ to unit $i$ has a scaling factor $M_{ij}(t)= \Lambda_i(t)/\Lambda_j(t)$. The $\lambda_i(t)$  must remain within a finite closed interval away from 0 and infinity. To see this, imagine for instance that there is a subsequence of $\Lambda_3(t)$ that approaches 0. Then there must be a corresponding subsequence of $\Lambda_4(t)$
   that approaches 0, or else the contribution of the weight $w_{43}\Lambda_4(t)/\Lambda_3(t)$ to the regularizer would go to infinity. But then, as we reach the output layer, the contribution of the last weight
   $w_{54}\Lambda_5(t)/\Lambda_4(t)$ to the regularizer goes to infinity because $\Lambda_5(t)$ is fixed to 1 and cannot compensate for the small values of $\Lambda_4(t)$. And similarly, if there is a subsequence of $\Lambda_3(t)$ going to infinity, we obtain a contradiction by propagating its effect towards the input layer. }  
			\label{fig:Limits}
\end{figure}

\begin{figure}[ht]
	\centering
		\includegraphics[keepaspectratio=true,width=1.0\textwidth]{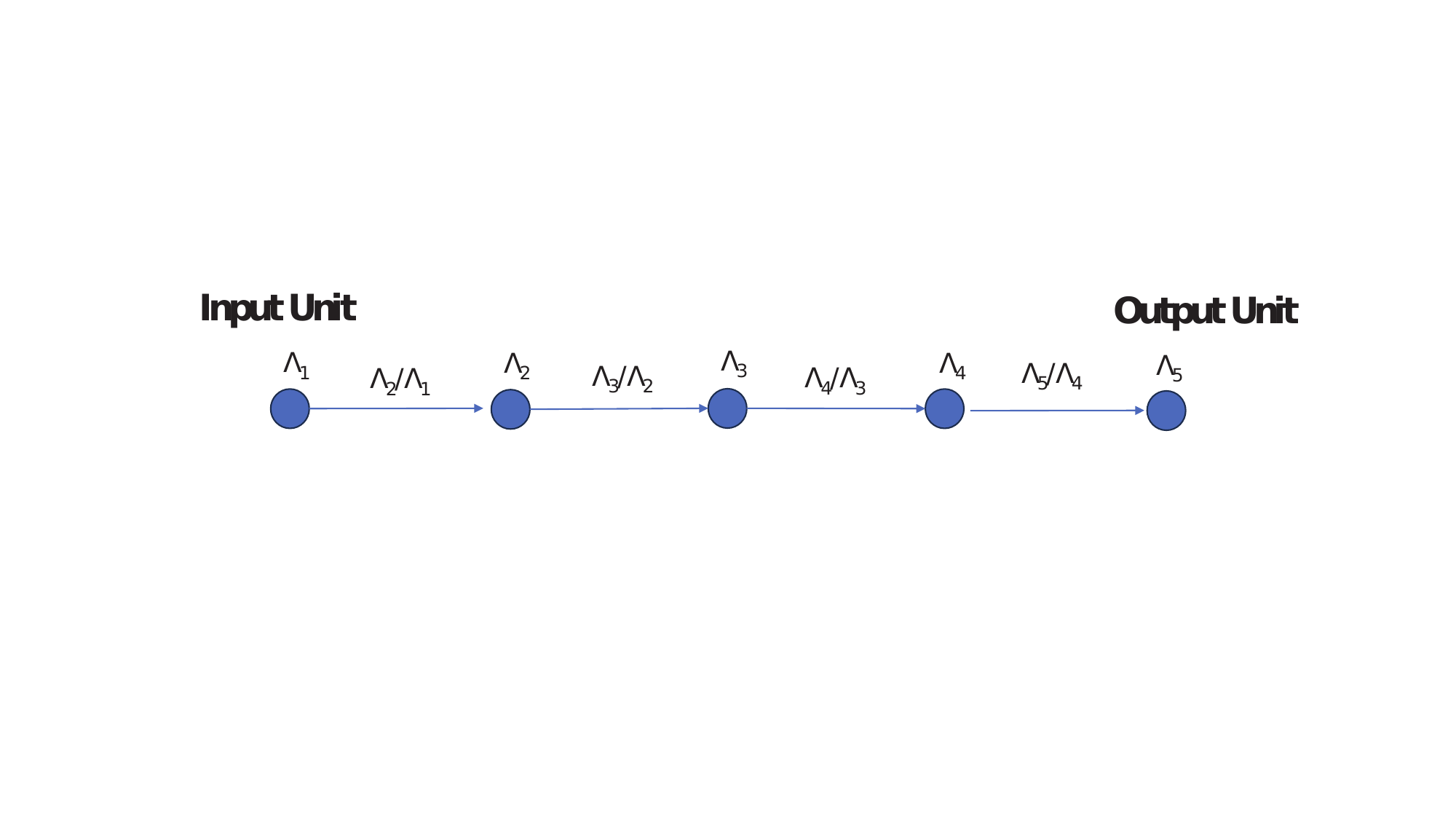}
  \vspace{-4cm}
		\caption[temp]
		{A path with five units. After the stochastic balancing algorithm has converged, each unit $i$ has a scaling factor $\Lambda_i$, and each directed edge from unit $j$ to unit $i$  has a scaling factor $M_{ij}= \Lambda_i/\Lambda_j$.   
  The products of the $M_{ij}$'s along the path is given by:
  $\frac{\Lambda_2}{\Lambda_1} \frac{\Lambda_3}{\Lambda_2} \frac{\Lambda_4}{\Lambda_3} \frac{\Lambda_5}{\Lambda_4}=\frac{\Lambda_5}{\Lambda_1}$. Accordingly, if we sum the variables  $L_{ij} = \log M_{ij}$ along the directed path, we get $L_{21}+L_{32}+L_{43}+L_{54}=\log \Lambda_5 - \log \Lambda_1$. In particular, if unit 1 is an input unit and unit 5 is an output unit, we must have $\Lambda_1=\Lambda_5=1$ and thus:  $L_{21}+L_{32}+L_{43}+L_{54}= 0$.  Likewise, in the case of a directed cycle where unit 1 and unit 5 are the same, we must have:  $L_{21}+L_{32}+L_{43}+L_{54}+ L_{15}= 0$.   }
				\label{fig:InputToOutput}
\end{figure}

\begin{figure}[ht]
	\centering
		\includegraphics[keepaspectratio=true,width=1.0\textwidth]{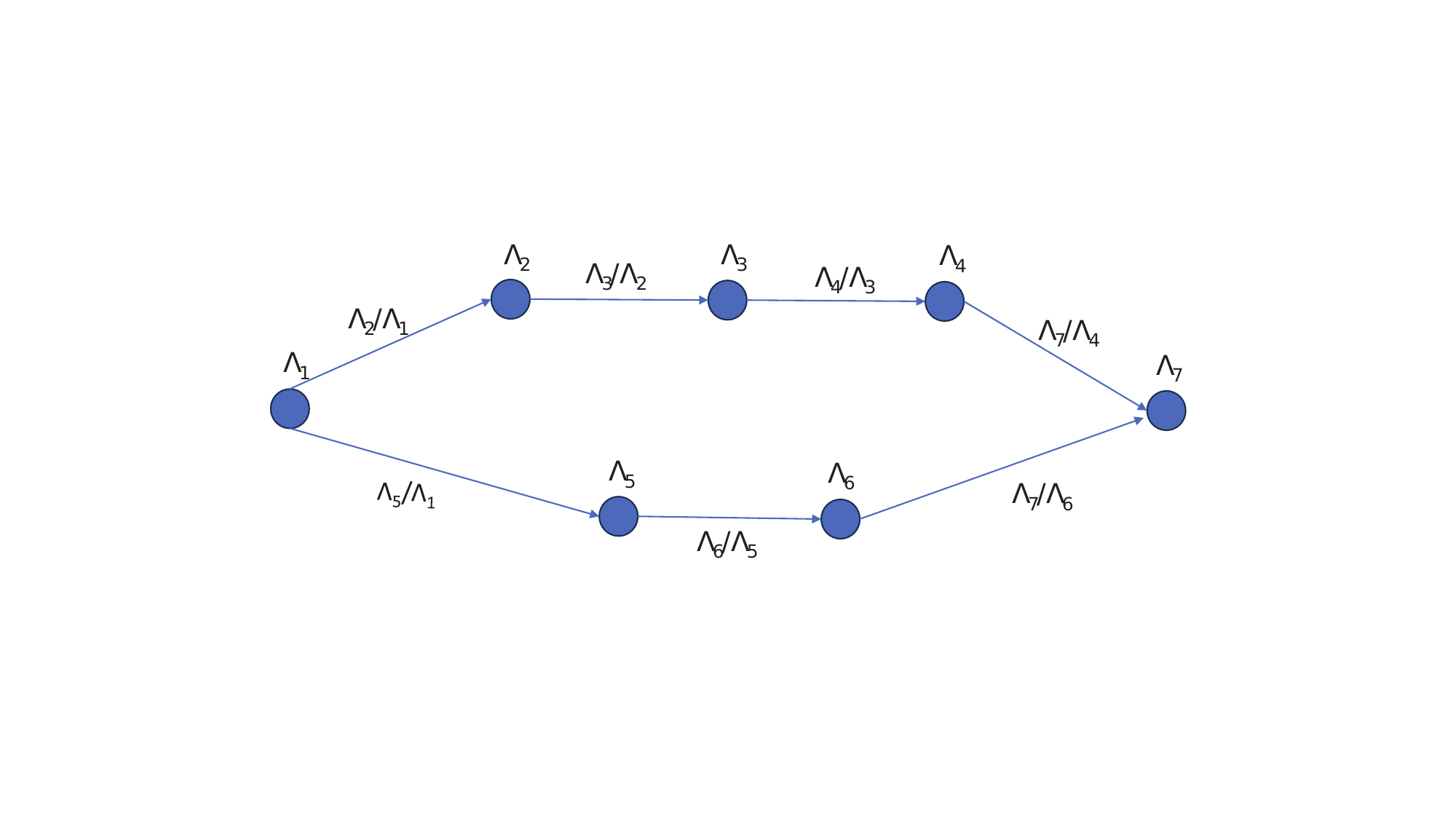}
  \vspace{-2cm}
		\caption[temp]
		{Two hidden units (1 and 7) connected by two different directed paths 1-2-3-4-7 and 1-5-6-7 in a BiLU network. Each unit $i$ has a scaling factor $\Lambda_i$, and each directed edge from unit $j$ to unit $i$  has a scaling factor $M_{ij}= \Lambda_i/\Lambda_j$. The products of the $M_{ij}$'s along each path is equal to:
  $\frac{\Lambda_2}{\Lambda_1} \frac{\Lambda_3}{\Lambda_2} \frac{\Lambda_4}{\Lambda_3} \frac{\Lambda_7}{\Lambda_4}=
  \frac{\Lambda_5}{\Lambda_1} \frac{\Lambda_6}{\Lambda_5} \frac{\Lambda_7}{\Lambda_6}=\frac{\Lambda_7}{\Lambda_1}$. Therefore the variables $L_{ij}=\log M_{ij}$ must satisfy the linear equation:  $L_{21}+L_{32}+L_{43}+L_{74}=L_{51}+L_{65}+L_{76}$
  =$\log \Lambda_7- \log \Lambda_1$.}  
					\label{fig:DoublePath}
\end{figure}

\begin{figure}[ht]
	\centering
		\includegraphics[keepaspectratio=true,width=1.0\textwidth]{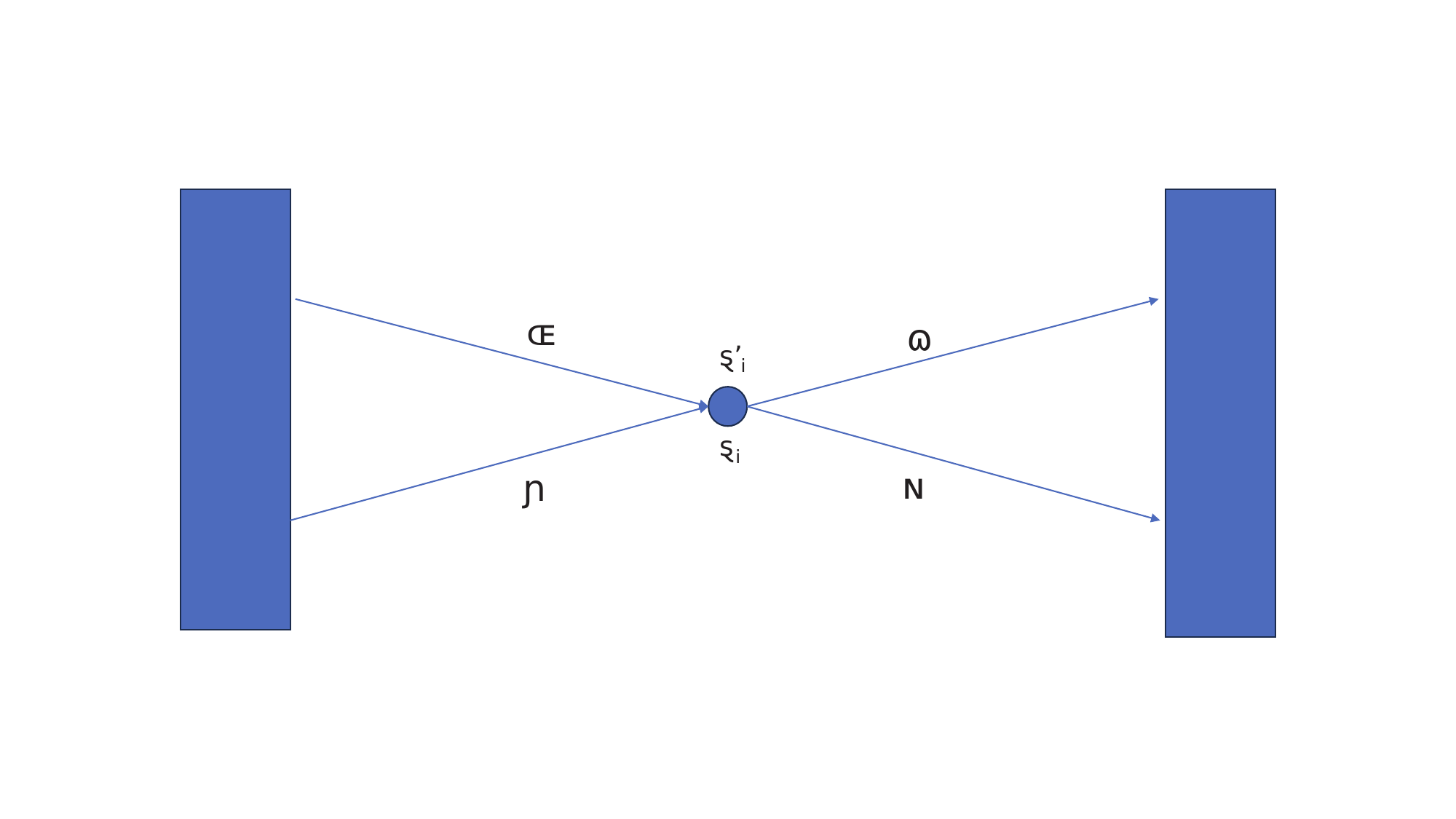}
  \vspace{-1cm}
		\caption[temp]
		{Consider two paths $\alpha+\beta$ and $\gamma + \delta$ from the input layer to the output layer going through the same unit $i$. Let us assume that the first path assigns a multiplier $\Lambda_i$ to unit $i$ and the second path assigns a multiplier $\Lambda'_i$ to the same unit.  By assumption we must have:
        $\sum_\alpha L_{ij} + \sum_\beta L_{ij}=0$ for the first path, and 
        $\sum_\gamma L_{ij} + \sum_\delta L_{ij}=0$. But $\alpha + \delta$ and $\gamma + \beta$ are also paths from the input layer to the output layer and therefore:
         $\sum_\alpha L_{ij} + \sum_\delta L_{ij}=0$ and  $\sum_\gamma L_{ij} + \sum_\beta L_{ij}=0$. As a result, $\sum_\alpha L_{ij}=\log \Lambda_i=\sum_\gamma L_{ij}=\Lambda'_i$. Therefore the assignment of the multiplier $\Lambda_i$ must be consistent across different paths going through unit $i$. }  
	\label{fig:TwoPaths}
\end{figure}

\begin{figure}[hpt!]
	\centering		\includegraphics[keepaspectratio=true,width=0.75\textwidth]{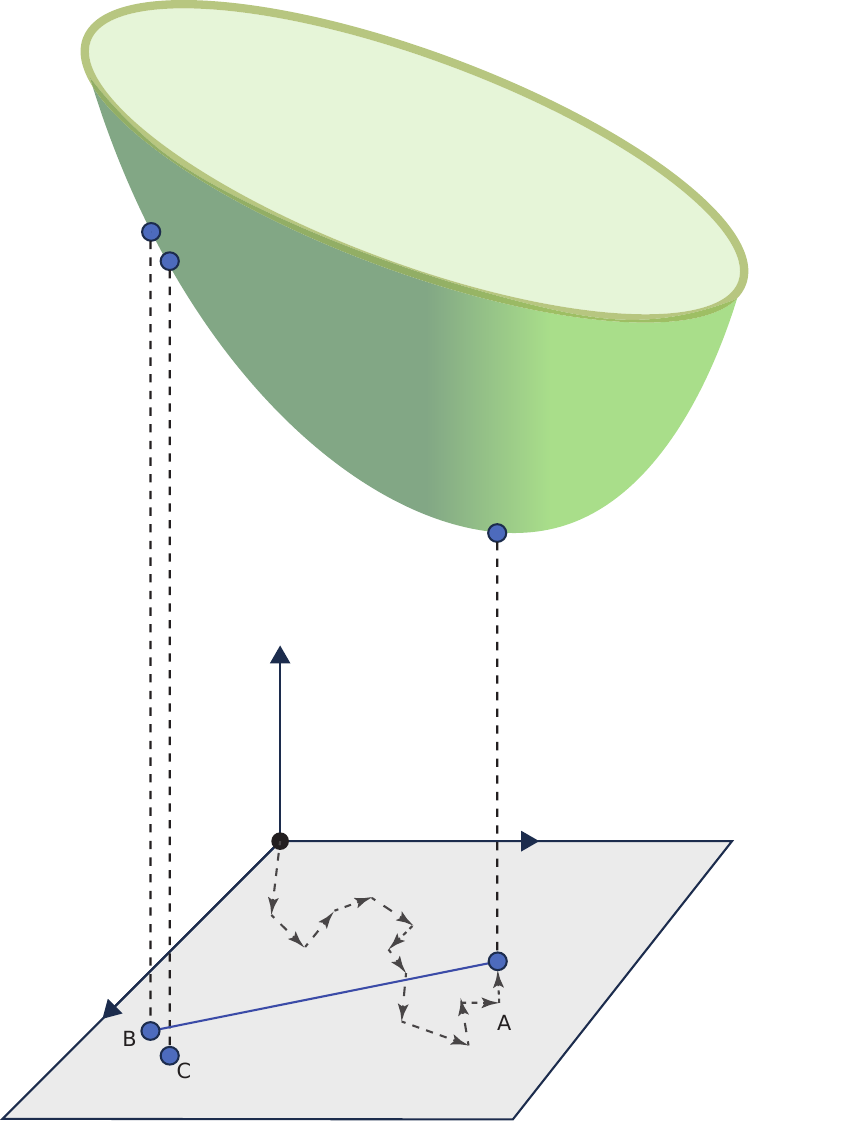}
  \vspace{0cm}
		\caption[temp]
		{ The problem of minimizing the strictly convex regularizer $R(L_{ij})=\sum_{ij} e^{pL_{ij}}\vert w_{ij} \vert^p$ ($p>0$),  over the linear (hence convex) manifold of self-consistent configurations defined by the linear constraints of the form $\sum_\pi L_{ij}=0$, where $\pi$ runs over input-output paths. The regularizer function depends on the weights. The linear manifold depends only on the architecture, i.e., the graph of connections. This is a strictly convex optimization problem with a unique solution associated with the point $A$. At $A$ the corresponding weights must be balanced, or else a self-consistent configuration of lower cost could be found by balancing any non-balanced neuron. Finally, any other self-consistent configuration $B$ cannot correspond to a balanced state of the network, since there must exist balancing moves that further reduce the regularizer cost (see main text). Stochastic balancing produces random paths from the origin, where $L_{ij=}\log M_{ij}=0$, to the unique optimum point $A$.}    	\label{fig:ConvexOpt}
\end{figure}

Finally, we prove that given an initial set of weights $W$, the final balanced state is unique and independent of the order of the balancing operations. The coefficients $\Lambda_i$ corresponding to a globally balanced state must be  solutions of the following optimization problem:

\be
\min_{\Lambda} R(\Lambda)= \sum_{ij} \vert \frac{\Lambda_i}{\Lambda_j} w_{ij} \vert^p
\label{eq:mini100}
\ee
under the simple constraints:
$\Lambda_i >0$ for all the BiLU hidden units, and 
$\Lambda_i=1$ for all the visible (input and output) units. 
In this form, the problem is not convex. Introducing new variables $M_j=1/\Lambda_j$ is not sufficient to render the problem convex. Using variables $M_{ij}=\Lambda_i/\Lambda_j$
is better, but still problematic for $0<p  \leq 1$.  However, let us instead introduce the new variables $L_{ij}=\log (\Lambda_i/\Lambda_j)$. These are well defined since we know that $\Lambda_i/\Lambda_j >0$. The objective now becomes:

\be
\min R(L)= \sum_{ij} \vert e^{L_{ij}} w_{ij} \vert^p=
\sum_{ij} e^{p L_{ij}} \vert w_{ij} \vert^p
\label{eq:mini200}
\ee
This objective is strictly convex in the variables $L_{ij}$, as a sum of strictly convex functions (exponentials). However, to show that it is a convex optimization problem we need to study the constraints on the variables $L_{ij}$. In particular, from the set of $\Lambda_i$'s it is easy to construct a unique set of $L_{ij}$. However what about the converse?

\begin{definition}
A set of real numbers $L_{ij}$, one per connection of a given neural architecture, is self-consistent if and only if there is a unique corresponding set of numbers $\Lambda_i>0$ (one per unit) such that: $\Lambda_i=1$ for all visible units and $L_{ij}=\log \Lambda_i/\Lambda_j$ for every directed connection from a unit $j$ to a unit $i$.
\end{definition}

\begin{remark}
This definition depends on the graph of connections, but not on the original values of the synaptic weights. 
Every balanced state is associated with a self-consistent set of $L_{ij}$, but not every self-consistent set of $L_{ij}$ is associated with a balanced state. 
\end{remark}

\begin{proposition}
A set $L_{ij}$  associated with a neural architecture is self-consistent if and only if 
$\sum_{\pi} L_{ij}=0$ where $\pi$ is any directed path connecting an input unit to an output unit or any directed cycle (for recurrent networks).
\end{proposition}

\begin{remark}
    Thus the constraints associated with being a self-consistent configuration of $ L_{ij}$' s are all linear. This resulting linear manifold $\mathcal L$  depends only on the architecture, i.e., the graph of connections, but not on the actual weight values. The strictly convex function $R (L_{ij})$ depends on the actual weights $W$. Different sets of weights $W$ produce different convex functions over the same linear manifold. If $E$ denotes the total number of connections, then obviously $\dim {\mathcal L} \leq E$. In order to infer all the $\Lambda_i$, there must exist at least one constrained path going through each node $i$. Thus, in a layered feedforward network, the dimension of $\mathcal L$ is given by:  $\dim {\mathcal L}=E-M$,  where here $M$ denotes the size of the largest layer. 
\end{remark}
\begin{remark}
  One could coalesce all the input units and all output units into a single unit, in which case a path from an input unit to and output unit becomes also a directed cycle. In this representation, the constraints are that the sum of the $L_{ij}$ must be zero along any directed cycle.  In general, it is not necessary to write a constraint for every path from input units to output units. It is sufficient to select a representative set of paths such that every unit appears in at least one path.   
\end{remark}

\begin{remark}
   All the results in this section remain true in a mixed network containing both BiLU neurons and non-BiLU neurons, as long as one uses $\Lambda_i=1$ for any non-BiLU neuron.    
\end{remark}

\begin{proof}
If we look at any directed path $\pi$ from unit $i$ to unit $j$, it is easy to see that we must have:

\be
\sum_\pi  L_{kl} = \log \Lambda_i -\log \Lambda_j
\label{eq:layer1000}
\ee
This is illustrated in Figures \ref{fig:InputToOutput} and \ref{fig:DoublePath}. 
Thus along any directed path that connects any input unit to any output unit, we must have $\sum_\pi L_{ij}=0$. In addition, for recurrent neural networks, if $\pi$ is a directed cycle we must also have:
 $\sum_\pi L_{ij}=0$. Thus in short we only need to add linear constraints of the form:
 $\sum_\pi L_{ij}=0$. Any unit is situated on a path from an input unit to an output unit. 
 Along that path, it is easy to assign a value $\Lambda_i$ to each unit by simple propagation starting from the input unit which has a multiplier equal to 1. When the propagation terminates in the output unit, it terminates consistently because the output unit has a multiplier equal to 1 and, by assumption, the sum of the multipliers along the path must be zero. So we can derive scaling values  $\Lambda_i$ from the variables $L_{ij}$. Finally, we need to show that there are no clashes, i.e. that it is not possible for two different propagation paths to assign different multiplier values to the same unit $i$. The reason for this is illustrated in Figure \ref{fig:TwoPaths}.
 \end{proof}

We can now complete the proof Theorem \ref{thm:uniqueness}. 
Given a neural network of BiLUs with a set of weights $W$, we can consider the problem of minimizing the regularizer $R(L_{ij}$ over the self-admissible configuration $L_{ij}$. For any $p>0$, the $L_p$ regularizer is strictly convex and the space of self-admissible configurations is linear and hence convex. Thus this is a strictly convex optimization problem that has a unique solution  (Figure \ref{fig:ConvexOpt}). Note that the minimization is carried over self-consistent configurations, which in general are not associated with balanced states. However, the configuration of the weights associated with the optimum set of $L_{ij}$ (point $A$ in Figure \ref{fig:ConvexOpt}) must be balanced. To see this, imagine that one of the BiLU units--unit $i$ in the network is not balanced.  Then we can balance it using a multiplier $\lambda_i^*$ and replace $\Lambda_i$ by 
$\Lambda_i'=\Lambda_i\lambda^*$. It is easy to check that the new configuration including $\Lambda'_i$ is self-consistent.  Thus, by balancing unit $i$, we are able to reach a new self-consistent configuration with a lower value of $R$ which contradicts the fact that we are at the global minimum of the strictly convex optimization problem. 

We know that the stochastic balancing algorithm always converges to a balanced state. We need to show that it cannot converge to any other balanced state, and in fact that the global optimum is the only balanced state. By contradiction, suppose it converges to a different balanced state associated with the coordinates $(L^B_{ij})$ (point $B$ in Figure \ref{fig:ConvexOpt}). Because of the self-consistency, this point is also associated with a unique set of $(\Lambda^B_i)$ coordinates. The cost function is continuous and differentiable in both the $L_{ij}$'s and the $\Lambda_i$'s coordinates. If we look at the negative gradient of the regularizer, it is non-zero and therefore it must have at least one non-zero component $\partial R/\partial \Lambda_i$ along one of the $\Lambda_i$ coordinates. This implies that by scaling the corresponding unit $i$ in the network, the regularizer can be further reduced, and by balancing unit $i$ the balancing algorithm will reach a new point ($C$ in Figure \ref{fig:ConvexOpt}) with lower regularizer cost. This contradicts the assumption that $B$ was associated with a balanced stated. 
Thus, given an initial set of weights $W$, the stochastic balancing algorithm must always converge to the same and unique optimal balanced state $W^*$ associated with the self-consistent point $A$.  A particular stochastic schedule corresponds to a random path within the linear manifold from the origin (at time zero all the multipliers are equal to 1, and therefore  for any $i$ and any $j$: $M_{ij}=1$ and $L_{ij}=0$) to the unique optimum point $A$.
\end{proof}

\begin{remark}
  From the proof, it is clear that the same result holds also for any deterministic balancing schedule, as well as for tied and non-tied subset balancing, e.g., for layer-wise balancing and tied layer-wise balancing. In the Appendix, we provide an analytical solution for the case of tied layer-wise balancing in a layered feed-forward network.  
\end{remark}

\begin{remark}
  The same convergence to the unique global optimum is observed if each neuron, when stochastically visited, is partially balanced (or favorably scaled) rather than fully balanced, i.e., it is scaled with a factor that reduces $R$ but not necessarily minimizes $R$, as long as these factors are not taken to be infinitesimally close to 1. .
  Stochastic balancing can also be viewed as a form of EM algorithm where the E and M steps can be taken fully or partially. Partial balancing can be faster and is used in some of the experiments of Section 
  \ref{sec:experiments}. 
  \end{remark}

\begin{remark}
Stochastic balancing of synapses can be applied to any network, even networks comprised entirely of non-homogeneous neurons. In this case, stochastic balancing will still converge to a unique stable configuration of the synaptic weights.  However, the overall function implemented by the balanced network may differ from the function implemented by the network at the start of the stochastic balancing. 
Experiments on balancing sigmoidal neurons are reported in Section \ref{sec:experiments}.
\end{remark}

\begin{remark}
In principle, balancing (or scaling) can be applied independently of the regularization approach being used. For instance, during training, one could alternate scaling and stochastic gradient steps, where the scaling step is performed with respect to $L_1$, but the gradient descent step is performed with respect to an $L_2$ regularized error function.  
\end{remark}

\begin{figure}[hpt]
	\centering		\includegraphics[keepaspectratio=true,width=0.85\textwidth]{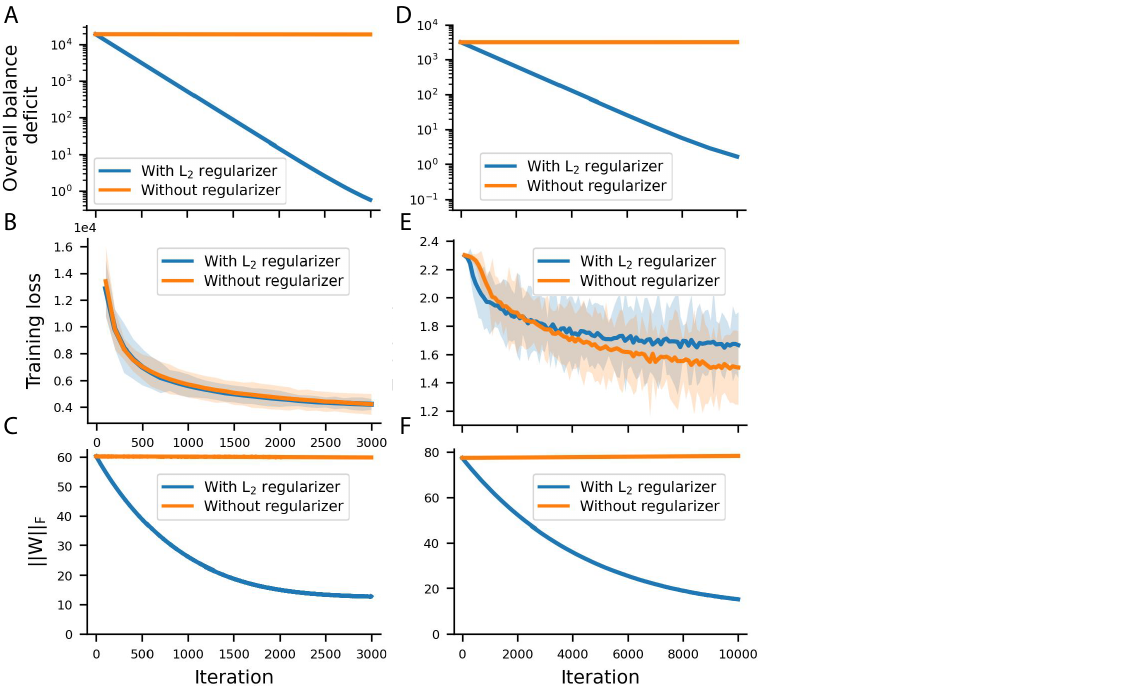}
		\caption[temp]
		{\textbf{SGD applied to $ E$ alone, in general, does not converge to a balanced state, but  sGD applied to $E+R$ converges to a balanced state.}  
		\textbf{(A-C)} Simulations use a deep fully connected autoencoder trained on the  MNIST dataset.
		\textbf{(D-F)} Simulations use  a deep locally connected network trained on the CIFAR10 dataset.
			\textbf{(A,D)} Regularization leads to neural balance.
   		\textbf{(B,E)} The training loss decreases and converges during training (these panels are not meant for assessing the quality of learning when using a regularizer).
     		\textbf{(C,F)}  Using weight regularization decreases the norm of weights. \textbf{(A-F)} Shaded areas correspond to one s.t.d around the mean (in some cases the s.t.d. is small and the shaded area is not visible).
		}
				\label{fig_regularizer_vs_no_regularizer}
\end{figure}

	\begin{figure}[hpt]
	\centering	\includegraphics[keepaspectratio=true,width=0.85\textwidth]{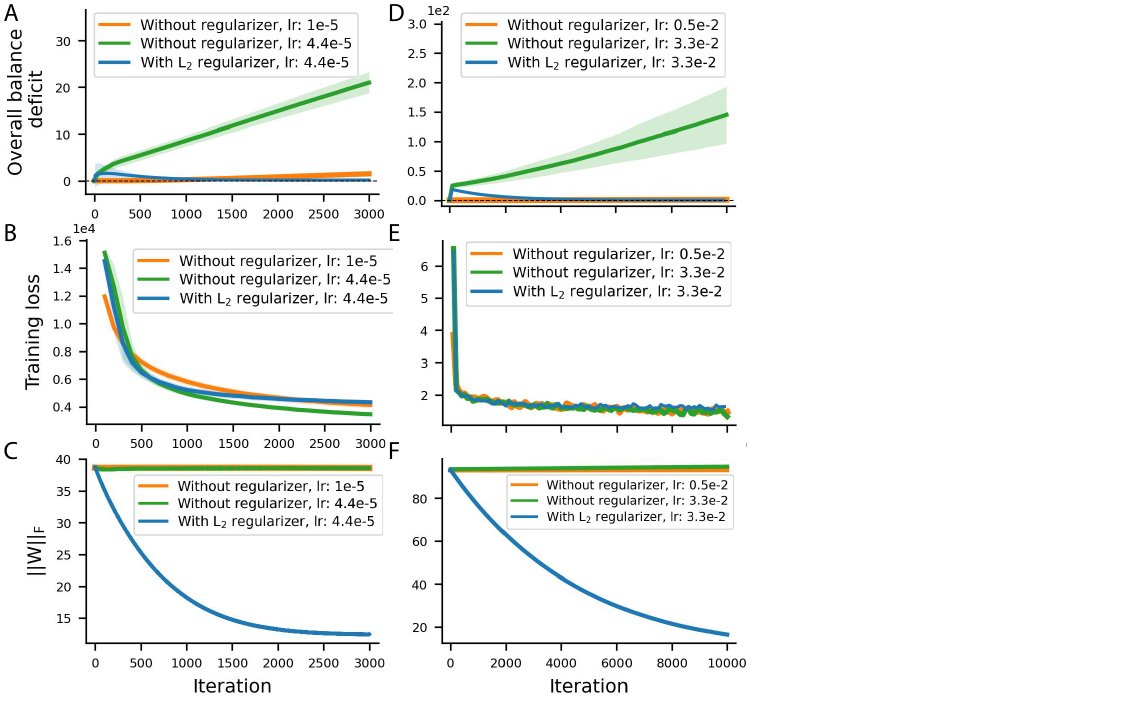}
	\caption[temp]
	{\textbf{Even if the starting state is balanced, SGD does not preserve the balance unless the learning rate is infinitely small. }    
	\textbf{(A-C)} Simulations use a deep fully connected autoencoder trained on the  MNIST dataset.
		\textbf{(D-F)} Simulations use  deep locally connected network trained on the CIFAR10 dataset.
 \textbf{(A-F)} The initial weights are balanced using the stochastic balancing algorithm. Then the network is trained by SGD. \textbf{(A,D)} When the learning rate (lr) is relatively large, without regularization, the initial balance of the network is rapidly disrupted. 
\textbf{(B,E)} 
The training loss decreases and converges during training (these panels are not meant for assessing the quality of learning when using a regularizer).
     		\textbf{(C,F)}  Using weight regularization decreases the norm of the weights. \textbf{(A-F)} 
       Shaded areas correspond to one s.t.d around the mean (in some cases the s.t.d. is small and the shaded area is not visible).    
       	}
    		\label{fig_initial_balance}
\end{figure}

\subsection{Convergence to a Unique Optimum for BiPU Stochastic Balancing }
We have seen that a generalized form of scaling and balancing can be defined for more general units than BiLUs, in particular for BiPUs.
Thus now we consider a network of units with activations functions $f$  satisfying the relationship: $f(\lambda x)= \lambda^cf(x)$ (note that this includes BiLU units for $c=1$). We even allow $c$ to vary from unit to unit. 

It is easy to see that most of the analyses above done for BiLU units apply to this generalization. In particular, if we apply stocahstic generalized balancing, in the limit the positive multipliers of each connection $w_{ij}$ must satisfy:

\be
M_{ij}= \Lambda_i / \Lambda_j^{c_j}
\label{eq:gen100}
\ee
As above, we can define a new set of variables $L_{ij}=\log M_{ij}$ and, for any $p>0$, the regularizer $R(L)=\sum_{ij} e^{pL_{ij}}\vert w_{ij} \vert^p$ is strictly convex. What is different, however, is the set of constraints on the variables $L_{ij}$. These are the constraints that allow one to compute the variables $\Lambda_i$ uniquely from the variables $L_{ij}$ (or, equivalently, the variables $M_{ij}$). This is addressed by the following theorem. 

\begin{theorem}
\label{thm:uniqueness10}
Under the same conditions of Theorem \ref{thm:uniqueness}, but using activation functions that satisfy for each unit $i$ the relationship
$f(\lambda x)= \lambda^{c_i}f(x)$, the corresponding stochastic  generalized balancing algorithm converges to the unique minimum of a 
strictly convex optimization problem in the variables $L_{ij}$. The strictly convex objective function is given by $R(L)=\sum_{ij} e^{pL_{ij}}\vert w_{ij} \vert^p$. The constraints are linear and of the form:

\be
\sum_{i \in \pi}  \left  (    \prod_{k=i} ^n  c_k       \right )           L_{ii-1}=0
\label{eq:gen105}
\ee
for each path $\pi$ from an input unit to an output unit, going sequentially through the units $0,1, \ldots, n$, where 0 corresponds to the input unit, and $n$ corresponds to the output unit of the path. The set of paths in the constraints must cover all the units in the network. 
\end{theorem}

\begin{proof}
Let us assume that there is a consistent set of multipliers $\Lambda_0, \ldots, \Lambda_n$ associated with the coefficients $L_{ii-1}=\log M_{ii-1}$ along the path $\pi$, with $\Lambda_0=\Lambda_n=1$.  
Since $M_{ii-1}= \Lambda_i/\Lambda_{i-1}^{c_{i-1}}$, we can derive the multipliers $\Lambda_i$ iteratively by propagating information from the input unit to the output unit, in the form:

\be
\Lambda_i= M_{ii-1} \Lambda_{i-1}^{c_{i-1}} \quad {\rm or} \quad 
\log \Lambda_i = L_{ii-1}+   c_{i-1}  \log \Lambda_{i-1}
\label{eq:gen110}
\ee
Using the boundary conditions $\Lambda_0=\Lambda_n=1$ gives the formula in Theorem \ref{thm:uniqueness10}. The same arguments given for BiLU units can be used to complete the proof. 
\end{proof}

\begin{remark}
Note that if all the units have the same exponent $c$ associated with the scaling of their activation functions, then the linear constraints have the simplified form: 

\be
\sum_{i \in \pi}  c^{n+1-i}     L_{ii-1}=0
\label{eq:gen115}
\ee
\end{remark}

Next we present two sets of simulations. The first set of simulations aims to corroborate the theory. The second set of simulations is more practical and shows how balancing can be used in practice during training to improve overall performance.  

\section{Simulations: Corroboration of the Theory}

The code and experiments for this section are publicly available at \url{https://github.com/ARahmansetayesh/a-theory-of-neural-synaptic-balance}.
To further corroborate the results, we ran multiple experiments. Here we report the results from two series of experiments. 
The first one is conducted using a six-layer, fully connected, autoencoder trained on MNIST \cite{deng2012mnist} for a reconstruction task with ReLU activation functions in all layers and the sum of squares errors loss function. The number of neurons in consecutive layers, from input to output, is  784, 200, 100, 50, 100, 200, 784. Stochastic gradient descent (SGD) learning by backpropagation is used for learning with a batch size of 200.

The second one is conducted using 
 three locally connected layers followed by three fully connected layers trained on CIFAR10 \cite{krizhevsky2009learning} for a classification task with leaky ReLU activation functions in the hidden layers, a softmax output layer, and the cross entropy loss function. The number of neurons in consecutive layers, from input to output, is 3072, 5000, 2592, 1296, 300, 100, 10. 
 Stochastic gradient descent (SGD) learning by backpropagation is used for learning with a batch size of 5.

\begin{figure}[H]%[hp!]
	\centering	\includegraphics[keepaspectratio=true,width=0.65\textwidth]{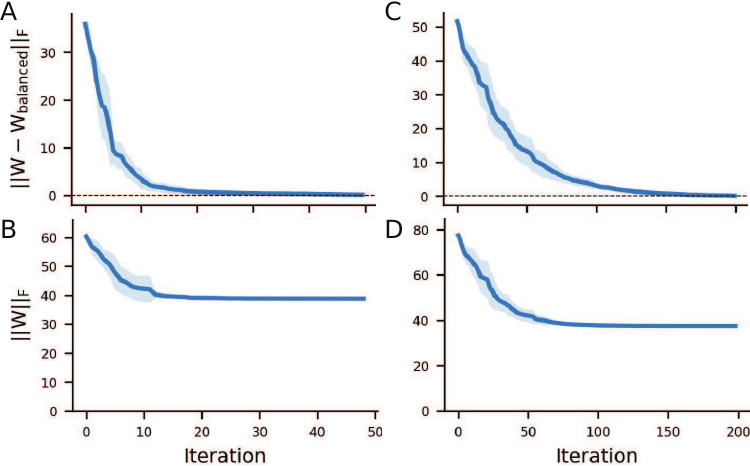}
	\caption[temp]
	{\textbf{Stochastic balancing converges to a unique global balanced state }
 \textbf{(A-B)} Simulations use a deep fully connected autoencoder trained on the  MNIST dataset.
		\textbf{(C-D)} Simulations use  deep locally connected network trained on the CIFAR10 dataset.
	 \textbf{(A,C)} The weights of the network are initialized randomly and saved. The stochastic balancing algorithm is applied and the resulting balanced weights are denoted by $W_{balanced}$. The stochastic balancing algorithm is applied 1,000 different times.  In all repetitions, the weights converge to the same value
	 $W_{balanced}$. 
		\textbf{(B,D)} Stochastic balancing decreases the norm of the weights.  
		\textbf{(A-D)} Shaded areas correspond to one standard deviation around the mean.
	}
 	\label{fig_stochastic_balancing}
\end{figure}

In all the simulation figures (Figures \ref{fig_regularizer_vs_no_regularizer}, \ref{fig_initial_balance}, and \ref{fig_stochastic_balancing}) the left column presents results obtained from the first experiment, while the right column presents  results obtained from the second experiment. While we  used both $L_1$ and $L_2$ regularizers in the experiments, in the figures we report the results obtained 
with the $L_2$ regularizer, which is the most widely used regularizer. In Figures \ref{fig_regularizer_vs_no_regularizer} and \ref{fig_initial_balance}, training is done using batch gradient descent on the MNIST and CIFAR data. The balance deficit for a single neuron $i$ is defined as:
$\bigl ( \sum_{w \in IN(i)} w^2 -\sum_{w\in OUT(i)}w^2 \bigr )^2
$,
and the overall balance deficit is defined as the sum of these single-neuron balance deficits across
all the hidden neurons in the network. The overall deficit is zero if and only if each neuron is in balance. In all the figures, $\vert \vert W \vert\vert_F$ denotes the Frobenius norm of the weights. 

Figure \ref{fig_regularizer_vs_no_regularizer} shows that learning by gradient descent with a $L_2$ regularizer results in a balanced state. Figure \ref{fig_initial_balance} shows that even when the network is initialized in a balanced state, without the regularizer the network can become unbalanced if the fixed learning rate is not very small. 
Figure \ref{fig_stochastic_balancing} shows that the local stochastic balancing algorithm, by which neurons are randomly balanced in asynchronous fashion, always converges to the same (unique)
global balanced state.

% \par\noindent {\bf Code Availability:}
% The code for reproducing the simulation results 
% is available under the Apache 2.0 license at:
% https://github.com/ARahmansetayesh/a-theory-of-neural-synaptic-balance.

\section{Simulations: Improving Training}
\label{sec:experiments}

The code and experiments for this section are publicly available at \url{
https://github.com/antonyalexos/Neural-Balance}.
In these simulations, we show that full balancing before training, as well as alternating partial balancing with stochastic gradient descent, are effective and practical regularization approaches that often outperform stochastic gradient descent paired with a traditional regularization function. 
We train all of our models on a server equipped with 8 Nvidia RTX A6000 Ada Generation graphics cards, with 384 GB of total memory, run on CUDA version 12.4. To ensure that our results are reproducible and fair, we repreat the experiments 8 times with 8 different random seeds shared across the different methodologies and take the average. In the experiments we use both $L_1$ and $L_2$ balancing. To assess both full balance before training and partial balance during training, we combine a variety of neural balancing operations. For the full balancing operation, we apply the balancing operation to the neurons of the model after weight initialization and before training until the ratio of the input to output norms converge to within 0.01 of 1. To assess partial balance during training, we perform partial balancing operations on the neurons of the model at every epoch during training. The partial balancing procedure applies the balancing operation to the neurons in the network only once, usually going from input neurons to output neurons without reaching equilibrium.  In all the Tables in this section, optimal results are shown in bold. 

\begin{figure}[H]
    \centering
    \begin{minipage}{0.5\textwidth}
        \centering
        \includegraphics[width=\textwidth, keepaspectratio]{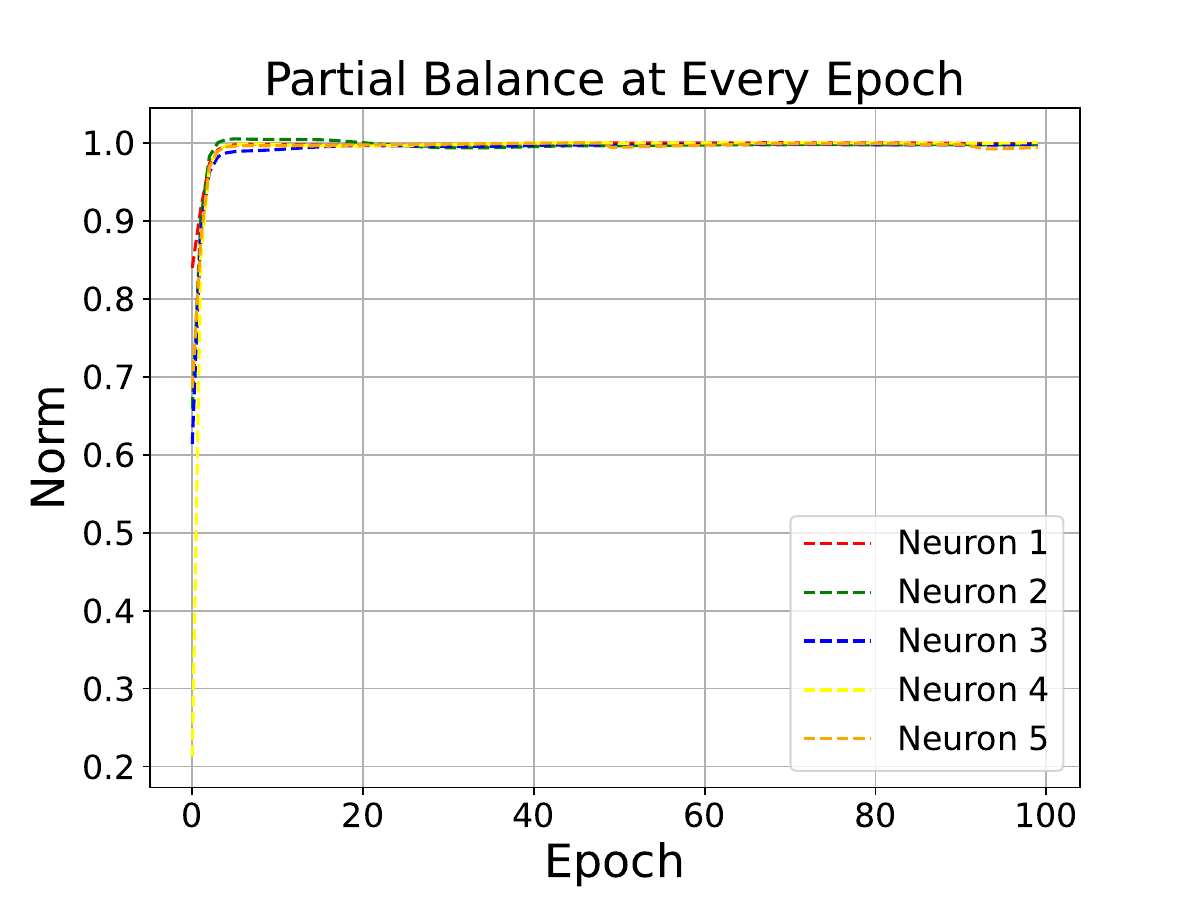}
        \caption{Partial neural balancing performed at every epoch on a toy model with 5 neurons, trained on a concentric-circle binary classification dataset. We observe that the norms of the model's weights gradually converge to 1.}
        \label{fig:toySemi} 
    \end{minipage}%
    \hfill
    \begin{minipage}{0.5\textwidth}
        \centering
        \includegraphics[width=\textwidth, keepaspectratio]{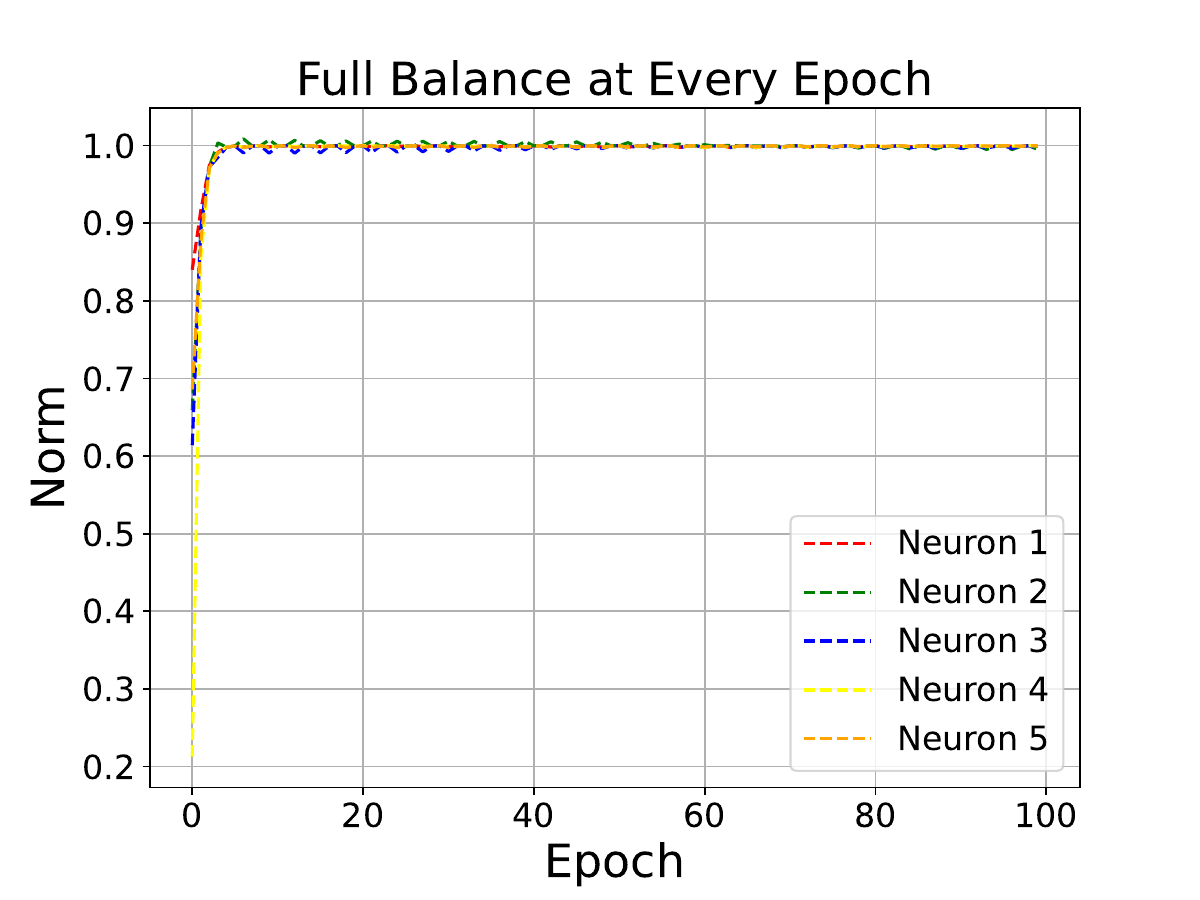}
        \caption{Full neural balancing performed at every epoch on a toy model with 5 neurons, trained on a concentric-circle binary classification dataset. We observe that the norms of the model's weights converge to 1 almost instantly.}
        \label{fig:toyFull}
    \end{minipage}
\end{figure}

\subsection{Toy Experiment on a Circle Toy Dataset}
\label{subsec:toyDataModel}

The toy experiments uses a simple 
 2-dimensional concentric circle clssification class and a simple network containing  2 input neurons, for each coordinate in the 2-dimensional plane, a hidden layer with 5 neurons, and a single output neuron for the binary classification task. The input and hidden layers apply a ReLU activation, and the output neuron uses a logistic activation for classification. The layers are fully connected. The loss is calculated using the binary cross entroyp for classification tasks. We use Adaptive Moment Estimation (Adam) as the optimizer during training, with a learning rate of $0.01$, and we train the model for 1000 epochs. 
To compare the outcome of full balancing with partial balancing operations, we perform both during training at every epoch. The full balancing procedure applies the balancing operation to the neurons in the network until the ratio of the input and output norms of every neuron converge within .01 of 1. The partial balancing procedure applies the balancing operation to the neurons of the neural network only once every epoch, proceeding from the input to the output neurons. 

Figure \ref{fig:toySemi} illustrates the effect of partial balancing applied at each learning epoch and shows that input and output weight norms become equalized for every neuron over time 
Figure \ref{fig:toyFull} illustrates the effect of full balancing with instantaneous equalization of the input and output norms of all the neurons. While both approaches converge to a balanced state, partial balancing is computationally less intensive and thus provides an effective alternative. 

\subsection{Full Balancing Before Training Applied to Fully Connected Networks}
\label{subsec:fullBalance}

In these experiments, we study the effect of full balancing prior to training on fully connected networks of various sizes. The networks are trained on the MNIST dataset. The networks are fully connected with an input size of 784, representing the flattened input shape of 28x28, and an output size of 10.
The 2 layer model has a single hidden layer with 256 neurons;  the 3 layer model has 2 hidden layers with   25 and 128 neurons; and the 5 layer model has 4 hidden layers with 512, 256, 128, and 64 neurons. All neurons have ReLU activation functions.  We optimize the cross entropy loss for classification with stochastic gradient descent with a learning rate of 0.001.
For experiments using $L_1$ and $L_2$ regularization, we use a regularization coefficient $\lambda = 0.015$ as we found it to be the most effective in this specific experiment via hyperparameter tuning.

\begin{figure}[H]
\centering
\includegraphics[width=0.85\textwidth, keepaspectratio]{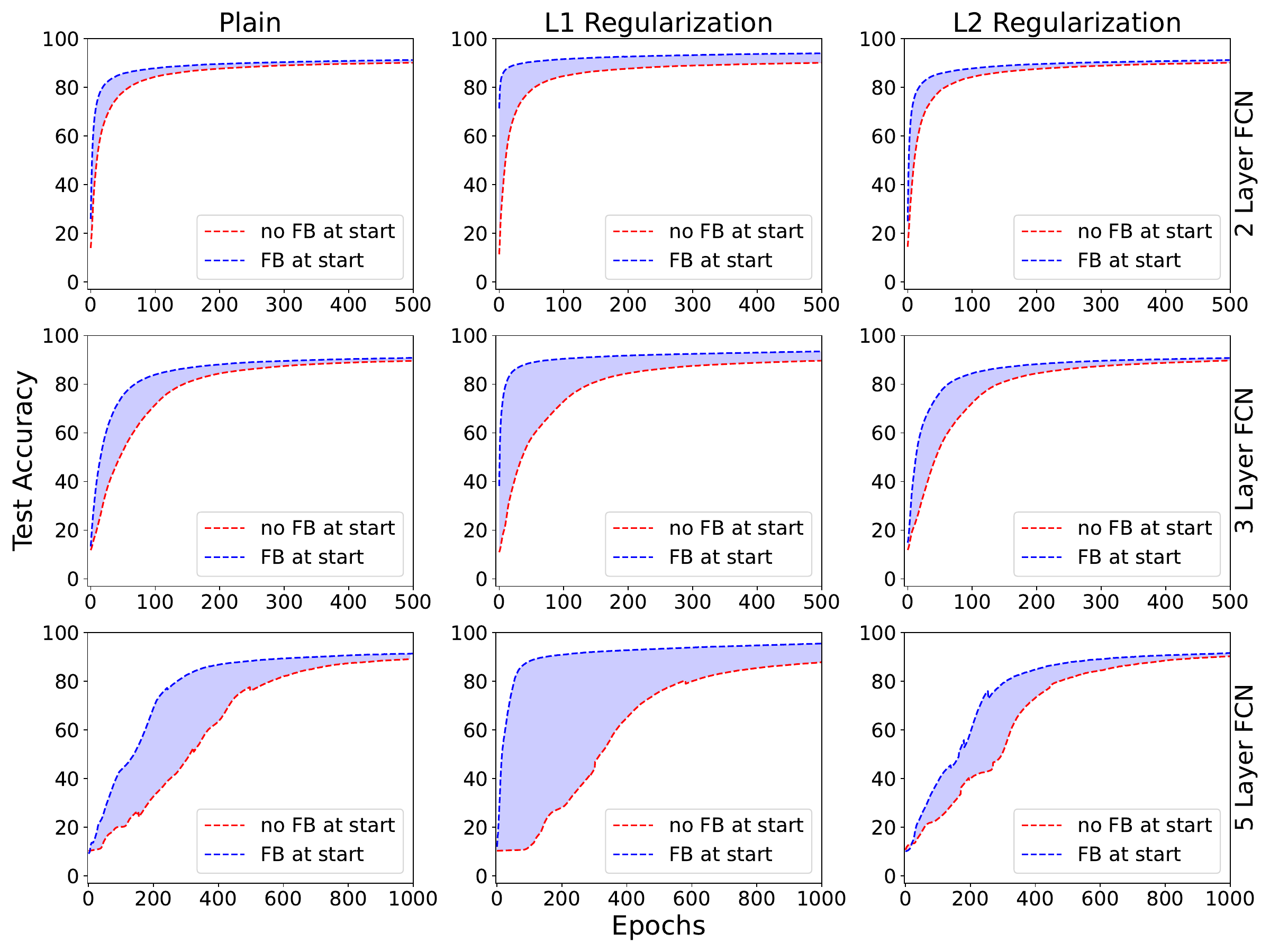}
\caption{The effect of full balancing (FB) before the start of training on various sizes of fully connected networks (FCN) with 2,3, and 5 hidden layers, trained on the MNIST dataset, without regularization, with $L_1$ regularization, and with $L_2$ regularization. FB at the beginning of learning results in faster convergence and higher test accuracy. The shading in each figure highlights the improvement. Table \ref{tab:noBalanceVsFullBalanceTable} reports the final text accuracy values.}
\label{fig:noBalanceVsFullBalance}
\end{figure}

In Figure \ref{fig:noBalanceVsFullBalance} and Table \ref{tab:noBalanceVsFullBalanceTable}, we evaluate the impact of applying the full balancing operation prior to the commencement of training. Compared to standard initialization, full balancing leads to faster convergence and higher overall accuracy when using the same model architecture, hyperparameters, and training methodologies. 
% The omission of partial balancing from the plots is due to its redundancy, as it inherently mirrors the effects of full balancing when performed at every epoch following an initial full balance. Repeated partial balancing yields the same outcome weights when using the same seed, although the weights converge over time since they are not balanced from the outset. 
Notably, the size of the benefit seems to increase with the size of the models. 
Figure \ref{fig:noBalanceVsFullBalance} provides a 
comparison between full balancing and no balancing applied to a Fully Connected Network (FCN) before training. Each square within the grid represents a combination of model size and training methodology, consistently demonstrating that neural balancing enhances the rate of convergence and model accuracy.

% \vspace{-6mm}

\begin{table}[H]
    \centering
    \resizebox{\textwidth}{!}{
        \begin{tabular}{|c|c|c|c|c|c|c|}
        \hline
        Type & \multicolumn{3}{|c|}{No FB at Start} & \multicolumn{3}{c|}{FB at Start} \\ \hline
        & Plain & L1 Reg. & L2 Reg. & Plain & L1 Reg. & L2 Reg.\\ \hline
        2 Layer FCN & 90.09\% & 90.05\% & 90.062\% & \textbf{91.22\%} & \textbf{93.96\%} & \textbf{91.18\%}
        \\ \hline
        3 Layer FCN & 89.594\% & 89.67\% & 89.70\% & \textbf{90.83\%} & \textbf{93.47\%} & \textbf{90.79\%}
        \\ \hline
        5 Layer FCN & 89.09\% & 87.85\% & 90.3\% & \textbf{91.37\%} & \textbf{95.50\%} & \textbf{91.59\%} \\
        \hline
        \end{tabular}
    }
    \caption{Test accuracy after training of plain, $L_1$-regularized, and $L_2$ regularized fully connected networks trained on MNIST, comparing full balancing (FB) before training with no balancing before training. As already observed in Figure \ref{fig:noBalanceVsFullBalance}, full balancing before training results in higher test accuracy.}
    \label{tab:noBalanceVsFullBalanceTable}
\end{table}

\subsection{Partial Balancing During Training Applied to  Fully Connected Networks}
\label{subsec:semibalancing-fcn_mnist}

Here we conduct the exact same experiments as in the previous section (\ref{subsec:fullBalance}),
but this time we use partial balancing during training in alternation with stochastic gradient descent steps.

\begin{figure}[H]
\centering
\includegraphics[width=1.\textwidth, keepaspectratio]{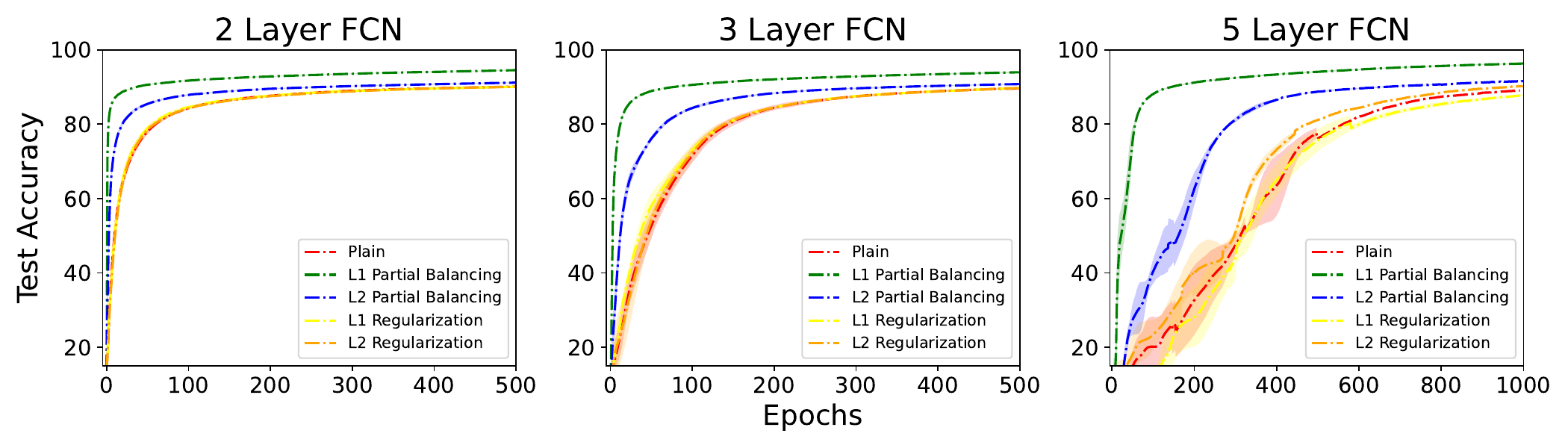}
\caption{
The effect of partial balancing during training on various sizes of fully connected networks (FCN) with 2,3, and 5 hidden layers, trained on the MNIST dataset, without regularization, with $L_1$ regularization, and with $L_2$ regularization. 
Partial balancing results in faster convergence and higher test accuracy. 
Table \ref{tab:ComparingMNISTPerformancesTable} contains the tabularized data from the plots for easier comparison.}
\label{fig:ComparingMNISTPerformancesPlot}
\end{figure}

From Figure \ref{fig:ComparingMNISTPerformancesPlot} and Table \ref{tab:ComparingMNISTPerformancesTable}, we observe that partial balancing during training  results in faster convergence and better test accuracy.

\begin{table}[H]
    \centering
    \resizebox{0.75\textwidth}{!}{
        \begin{tabular}{|l|c|c|c|c|c|}
        \hline
        Type & Plain & L1 PB & L2 PB & L1 Reg. & L2 Reg.\\
        \hline
        2-FCN & 91.22\% & \textbf{94.54\%} & 91.19\% & 93.96\% & 91.18\% \\
        \hline
        3-FCN & 90.84\% & \textbf{93.94\%} & 90.86\% & 93.47\% & 90.79\% \\
        \hline
        5-FCN & 91.37\% & \textbf{96.26\%} & 91.63\% & 95.48\% & 91.59\% \\
        \hline
        \end{tabular}
    }
    \caption{Comparison of test accuracy obtained with different training methods and 3 architectures of increasing depth size. 
 As already observed in Figure \ref{fig:ComparingMNISTPerformancesPlot},
 L1 partial balancing outperforms the other training methodologies on all model sizes.}
    \label{tab:ComparingMNISTPerformancesTable}
\end{table}

For increased robustness, we conduct the same tests using the FashionMNIST dataset. 
We utilize similar fully connected networks  of various sizes and implement partial balancing at every epoch.
Similar improvements in convergence and performance are observed. Irrespective of model size or training methodology, partial neural balancing markedly enhances the rate of convergence and overall test accuracy. As the models grow bigger, the inclusion of partial neural balancing in training helps models converge faster and perform better when compared to the other techniques.

\begin{figure}[H]
\centering
\includegraphics[width=1.\textwidth, keepaspectratio]{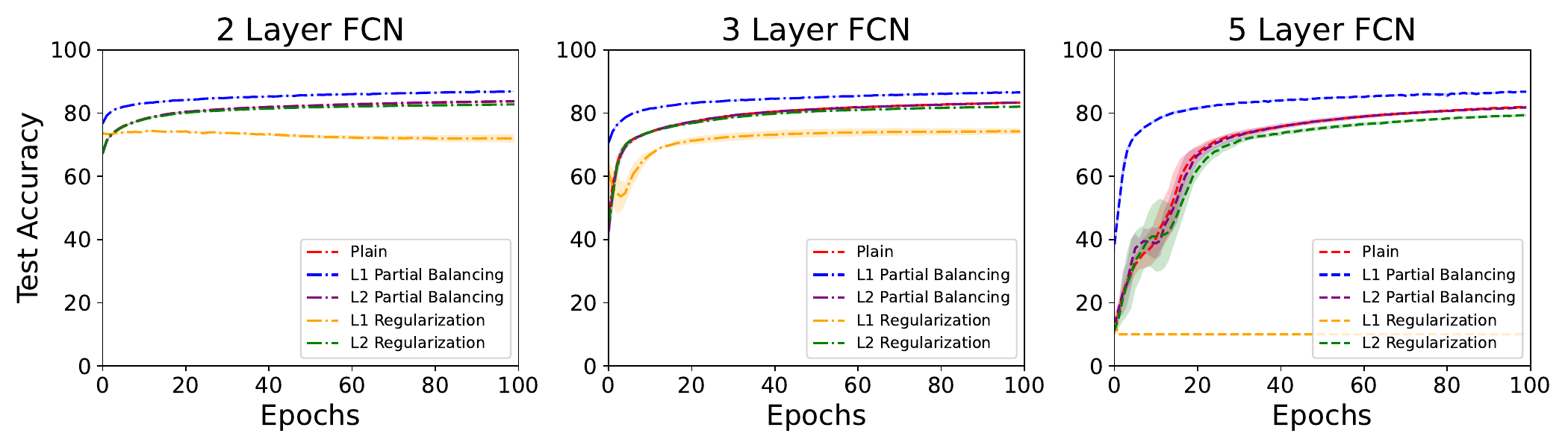}
\caption{
The effect of partial balancing during training on various sizes of fully connected networks (FCN) with 2,3, and 5 hidden layers, trained on the FashionMNIST dataset.
Partial balancing results in faster convergence and higher test accuracy. As the size of the models increases, partial balancing during training helps models converge faster and perform better.}  
\label{fig:ComparingFMNISTPerformancesPlot}
\end{figure}

% \subsection{Full Balance on Fully Connected Networks}
% \label{subsec:neural-balance-FCN}

% \begin{figure}[H]
% \centering
% \includegraphics[width=.8\textwidth, keepaspectratio]{experiment_figures/noBalanceVsFullBalanceTest.pdf}
% \caption{A comparison, using various training methodologies, between no balancing, and full balancing before training. We observe that full balancing positively impacts performance regardless of the model size or specific training methodology.}
% \label{fig:noBalanceVsFullBalance_full_balance}
% \end{figure}

% In this experiment we use 3 different FCNs of varying length (2,3,5 layers) and differnt training schemes, such as L1 and L2 regularization, 

% We continue our assessment of neural balancing with experiments performed on the RNN architecture. We train a 3-layered RNN on the IMDB sentiment analysis dataset, once again assessing full neural balancing with a 'plain', and regularized models. Figure \ref{fig:Rnn3Layers} demonstrates that when full balancing is performed before training, the model has a better final accuracy when compared to equivalent, non-balanced methodologies.

\subsection{Full Balancing Before Training Applied to Recurrent Neural Networks}
\label{subsec:neural-balance-rnn}

We continue the assessment of synaptic neural balancing by applying it to recurrent neural networks (RNNS). More specifically we assess the effect of full balancing before training. The experiments are carried on the  IMDB sentiment analysis dataset. We use a recurrent neural network with
an input layer, and output layer, and a fully-connected recurrent hidden layer. Plain text is first passed through an embedding layer, converting it to an embedding vector of length 100. The network uses 256 hidden neurons, which are fully connected among themselves. These recurrent neurons are updated three times synchronously before computing the final output. The final classification output is produced by a single logistic neuron. We use the binary cross entropy loss for training together with stochastic gradient descent with a learning rate of 0.001.
For experiments using $L_1$ and $L_2$ regularization, we use a regularization rate of $\lambda = 0.0001$ as we found it to be the most effective via hyperparameter tuning.

The results are shown in Table \ref{tab:Rnn3LayersTable}, demonstrating that, in all cases, when full balancing is performed before training, the model has better final accuracy.

% \begin{figure}[H]
% \centering
% \includegraphics[width=1.0\textwidth, keepaspectratio]{experiment_figures/RNN3Layers_journal.pdf}
% \caption{A comparison between partial balancing, L2 Regularization, and Plain Accuracy on a 3 Layer RNN using the IMDB sentiment analysis dataset. We also contrast the standard initialization with a full neural balancing operation performed before the start of training. We observe that neural partial balancing performed every epoch, paired with a full balance before training, results in the best overall accuracy, and convergence speed.}
% \label{fig:Rnn3Layers}
% \end{figure}

\begin{table}[H]
    \centering
    \resizebox{0.7\textwidth}{!}{
        \begin{tabular}{|l|c|c|c|}
        \hline
        Type & Plain & L1 Regularization & L2 Regularization\\
        \hline
        No FB at Start & 88.26 & 88.23 & 88.22 \\
        \hline
        FB at Start & \textbf{88.64\%} & \textbf{88.24\%} & \textbf{88.57\%} \\
        \hline
        \end{tabular}
    }
    \caption{Test accuracy for a recurrent neural network trained on the IMDB sentiment analysis dataset, comparing plain, $L_1$-regularized, and $L_2$-regularized models with and without full balancing (FB) at the start of training. In all cases, full balancing before training results in higher test accuracy at the end of training.}
    \label{tab:Rnn3LayersTable}
\end{table}

\subsection{Partial Balancing During Training Applied to Recurrent Neural Networks}
\label{subsec:semibalancing-rnn_imdb}

Similarly, here we assess the application of partial balancing during training, and demonstrate its efficacy at increasing the speed of convergence and the overall test accuracy, for recurrent neural network architectures. The experiments are also carried out on the IMDB sentiment analysis dataset, using the same recurrent network as in the previous section. 

\begin{figure}[H]
\centering
\includegraphics[width=1.\textwidth, keepaspectratio]{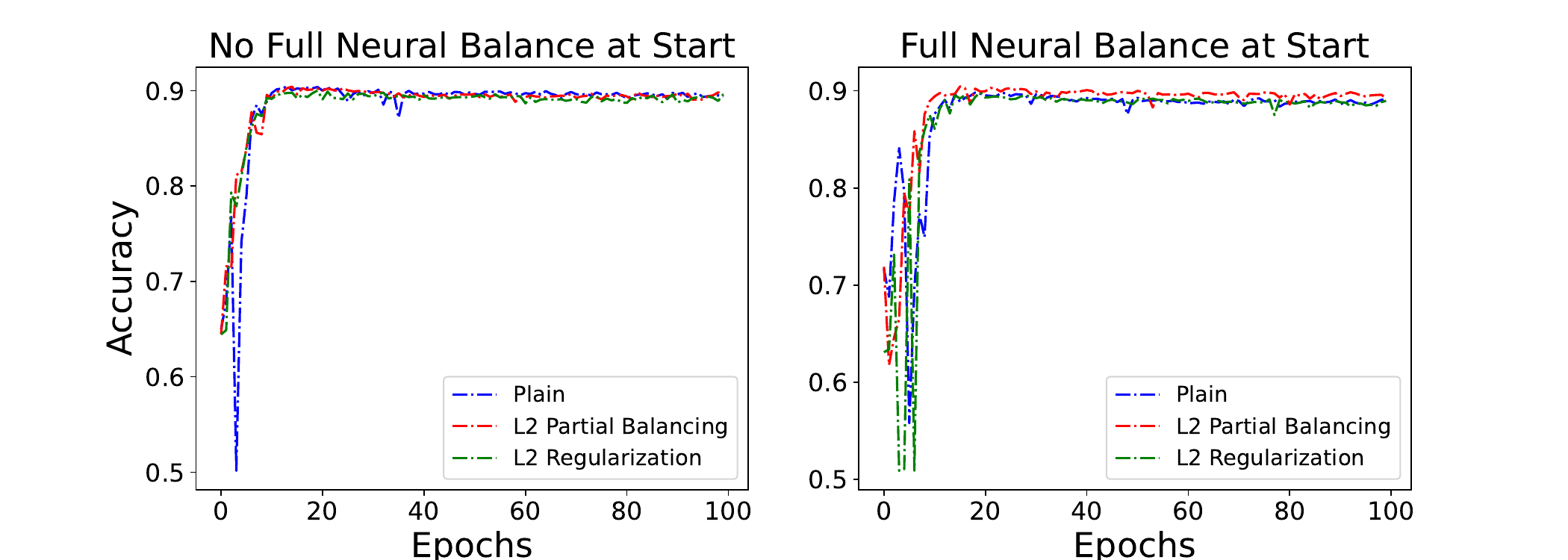}
\caption{Comparison between plain training (no regularization, no partial balancing) training with partial balancing, and training with $L_2$ regularization of a recurrent neural network on the IMDB dataset. We also contrast the standard random initialization with full balancing performed before the start of training.  Both partial balancing during training, and full balancing before training, result in faster convergence, and higher overall accuracy.}
\label{fig:Rnn3Layers_partial}
\end{figure}

The main results are summarized in Figure \ref{fig:Rnn3Layers_partial} where we compare the test accuracy achieved with plain training (no regularization and no balancing), training with partial balancing, and training with $L_2$ regularization.
We also compare implementing full balancing  before the start of training to standard random initialization. Both partial balancing during training and full balancing before training lead to faster convergence and higher accuracy.  

\subsection{Neural Balancing with Limited Data Environments}
\label{subsec:limited-data}

To further examine the reegularizing benefits of synaptic balancing, here we vary the amount of available training data.
For these experiments, we use both the MNIST handwritten digit classification dataset and the IMDB sentiment analysis dataset. Unlike the previous experiments, we use a stratified sample of the dataset to simulate a limited-data environment for the models. We use Scikit-Learn to take a stratified split of the overall dataset. For the MNIST dataset, we take 600 samples, representing 1\% of the full dataset, with 60 samples from each label, in order to maintain the label balance present in the main dataset. For the IMDB dataset, we take 1250 samples, representing 5\% of the full dataset, with 625 samples from the positive and negative labels, to preserve an equal distribution of labels.

For the experiments with the MNIST handwritten digit recognition dataset, we use the same fully connected network architecture used for the experiments that are performed on the whole dataset. We use the cross entropy loss with stochastic gradient descent with a learning rate of 0.001.
For experiments using $L_1$ and $L_2$ regularization, we use a regularization rate $\lambda = 0.015$ as we found it to be the most effective via hyperparameter tuning.
We test a combination of full balancing and partial balancing before or during training. With full balancing, the balancing operation is performed after initialization and before training until the ratio of the input to output norms of each neuron equalizes within a threshold of .001 of 1. For partial balancing, we balance each neuron once from inputs to output. This partial balancing step is carried out at the end of each training epoch.

For the experiments with the IMDB sentiment analysis dataset, we use the same recurrent neural network architecture that is used for the experiments performed on the whole dataset. 
We use binary cross entropy loss with stochastic gradient descent with a learning rate of ).001.
 For experiments using $L_2$ regularization, we use a regularization rate $\lambda = 0.0001$ as we found it to be the most effective via hyperparameter tuning.
We assess the performance of partial balancing by comparing it to $L_2$ regularization and plain train (no regularizaation, no balancing). 
For partial balancing, we balance each neuron once from inputs to output. This partial balancing step is carried out at the end of each training epoch.

\begin{figure}[H]
    \centering
    \includegraphics[width=.85\textwidth, keepaspectratio]{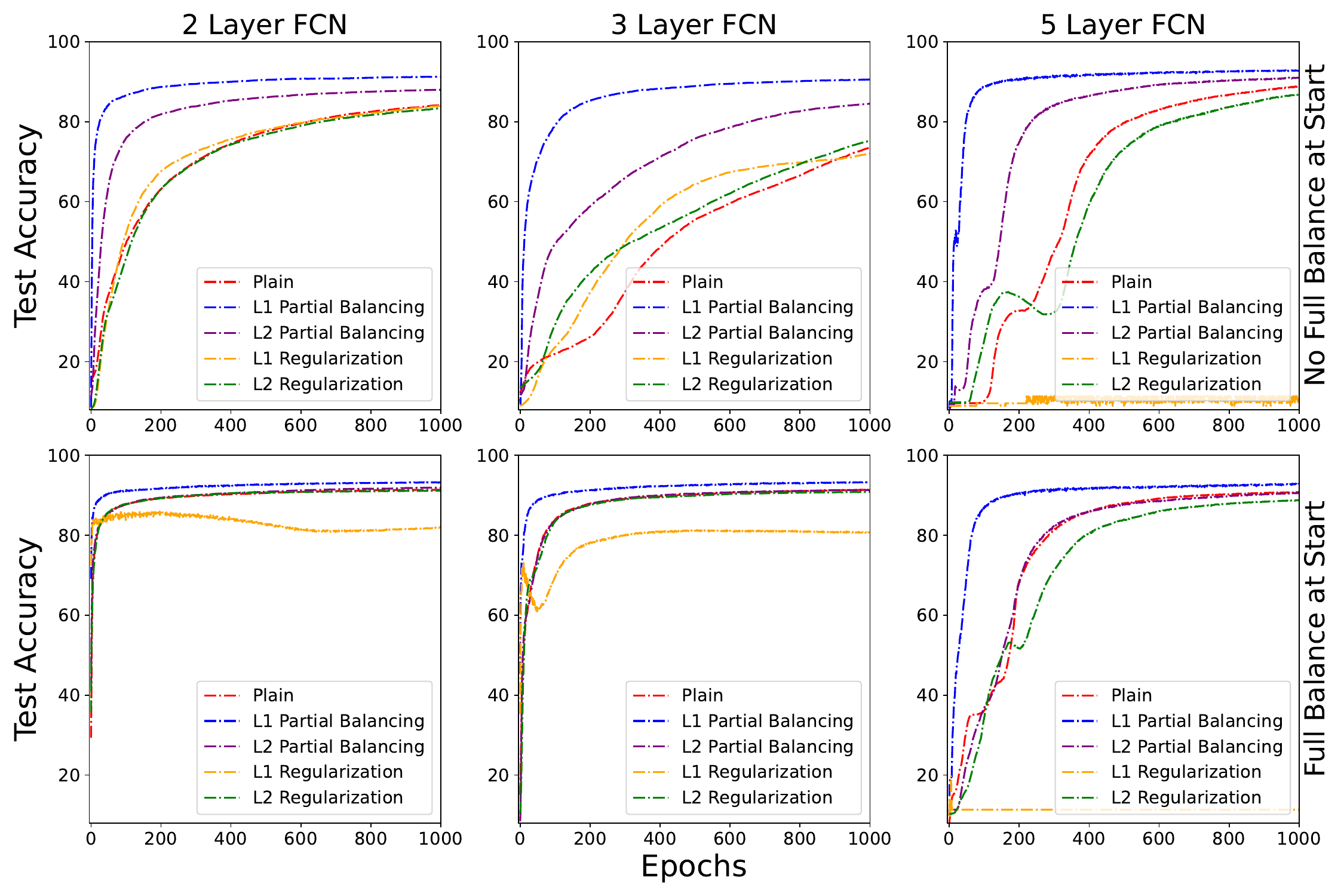}
    \caption{Comparison between plain training (no balancing, no regularization), training with partial balancinge, and training with regularization, with and without a full balancing at the start of training, using different fully connected networks trained on only 1\% of the MNIST dataset. Table \ref{tab:MnistFrac1e-2Table} shows the final test accuracy results for each of the  training methodologies. Nneural balancing consistently improves both the rate of convergence and the overall accuracy of the model.}
    \label{fig:MnistFrac1e-2}
\end{figure}

\begin{table}[H]
    \centering
    \resizebox{1.0\textwidth}{!}{
        \begin{tabular}{|c|c|c|c|c|c|c|}
        \hline
         Type & \multicolumn{3}{|c|}{No FB at Start} & \multicolumn{3}{c|}{FB at Start} \\ \hline
         & 2 Layer FCN & 3 Layer FCN & 5 Layer FCN & 2 Layer FCN & 3 Layer FCN & 5 Layer FCN\\ \hline
        Plain & 84.15\% & 73.49\% & 88.9\% & \textbf{91.39\%} & \textbf{91.42\%} & \textbf{90.86\%}
        \\ \hline
        L1 PB & 91.25\% & 90.57\% & 92.87\% & \textbf{93.26\%} & \textbf{93.3\%} & \textbf{92.92\%}
        \\ \hline
        L2 PB & 87.99\% & 84.53\% & 91.06\% & \textbf{91.94\%} & \textbf{91.37\%} & \textbf{90.59\%} \\
        \hline
        L1 Reg. & 83.92\% & 72.03\% & 11.35\% & \textbf{85.98\%} & \textbf{81.33\%} & \textbf{19.79\%} \\
        \hline
        L2 Reg. & 83.35\% & 75.21\% & 86.81\% & \textbf{91.16\%} & \textbf{90.86\%} & \textbf{88.78\%} \\
        \hline
        \end{tabular}
    }
    \caption{Test accuracy of various methodologies trained using 1\% of the MNIST dataset to simulate a limited data environment. Consistently with  Figure \ref{fig:MnistFrac1e-2}, in all cases the use of full balancing at the start of training not only increases the rate of convergence, but also leads to higher accuracy.}
    \label{tab:MnistFrac1e-2Table}
\end{table}

In data-scarce environments, neural networks incorporating balancing techniques exhibit accelerated convergence and enhanced accuracy compared to both unmodified models and those employing traditional regularization methods. Both full balancing at the start of learning, and partial balancing during learning lead to notable improvements.

\begin{figure}[H]
    \centering
    \includegraphics[width=0.7\textwidth, keepaspectratio]{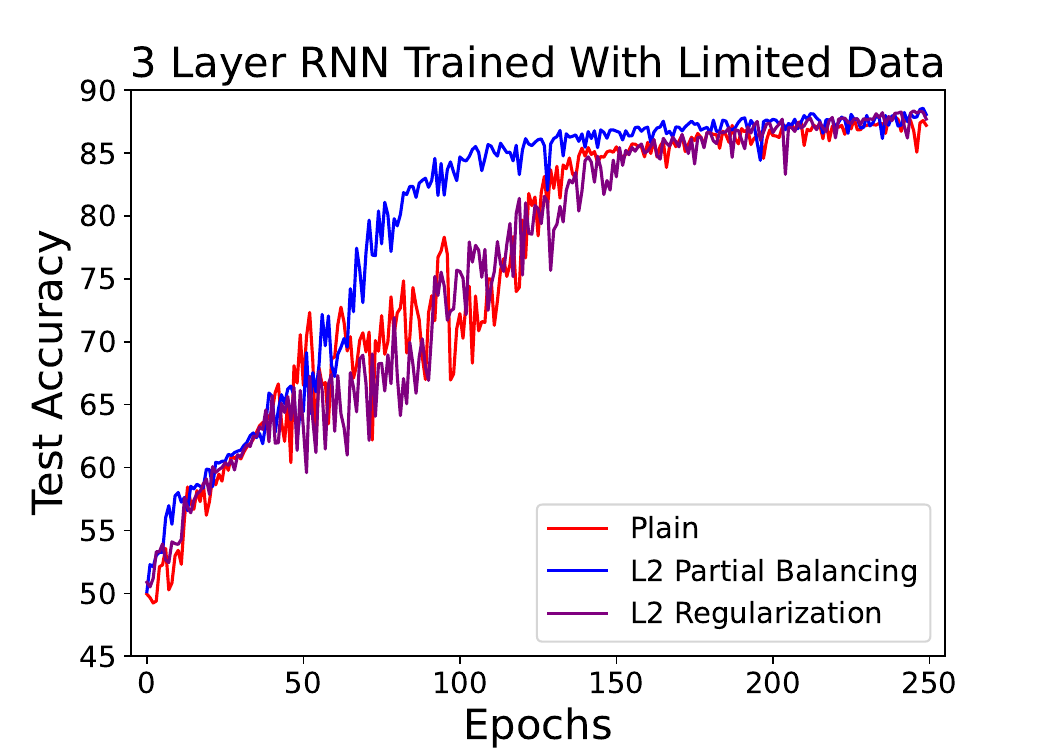}
    \caption{Comparison between plain training (no regularization, no balancing), training with partial balancing, and training with $L_2$ regularization using a RNN and only 5\% of the IMBD dataset. Partial balancing leads to the best overall performance. Table \ref{tab:05IMDBRNNTable} shows the final test accuracy. }
    \label{fig:05IMDBRNN}
\end{figure}

\begin{table}[H]
    \centering
    \resizebox{0.8\textwidth}{!}{
        \begin{tabular}{|l|c|c|c|}
        \hline
        Type & Plain & L2 Partial Balancing & L2 Regularization\\
        \hline
        &87.87\% & \textbf{88.55\%} & 88.34\% \\
        \hline
        \end{tabular}
    }
    \caption{Test accuracy for a recurrent neural network trained on 5\% of the IMDB sentiment analysis dataset, comparing plain training, training with partial balancing, and training with $L_2$ regularization. Consistently with the results in Figure \ref{fig:05IMDBRNN}, $L_2$ partial balancing performed at every epoch leads to the highest degree of accuracy. }
    \label{tab:05IMDBRNNTable}
\end{table}

We continue this assessment by training the same RNN as above using only 5\%  of the IMBD dataset. The results are provided in Figure \ref{fig:05IMDBRNN} and Table \ref{tab:05IMDBRNNTable}. Partial balancing leads to  faster convergence and higher accuracy, and thus in general balancing can improve generalization  in data scarce environments.

\subsection{Neural Balance with Non-BiLU Activation functions}
\label{subsec:tanh}
In all the previous experiments, we used neurons with ReLU activation functions. Here, we conduct similar experiments with non-BiLU neurons, in particular using sigmoidal neuron ($\tanh$ and logistic). For these experiments, we parallel the full balance experiments in section \ref{subsec:fullBalance}, and the partial balancing experiments in section \ref{subsec:semibalancing-fcn_mnist}. For both experiments, we use the same fully connected fedforward architectures with the same sizes, and the same hyperparameter settings as in the aforementioned sections.

\begin{figure}[H]
\centering
\includegraphics[width=0.9\textwidth, keepaspectratio]{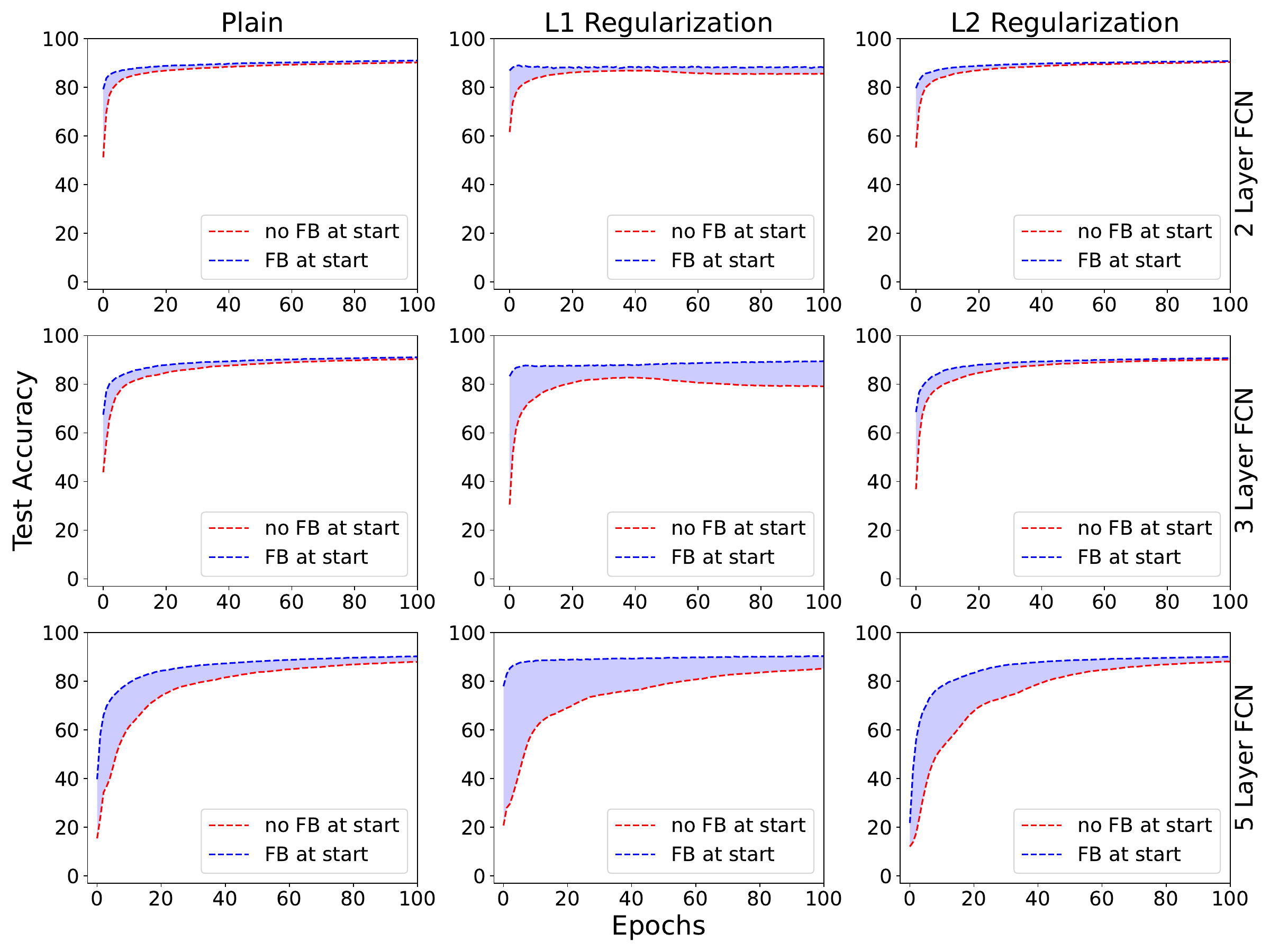}
\caption{Effect of full balancing (FB) before the start of training of fully connected feedforward networks of various depths, with $\tanh$ activation functions, using different training approaches: plain (no regularization, no balancing), training with $L_1$ or $L_2$ regularization. Regardless of the training method, full balancing at the start results in faster convergence and higher accuracy. The shading illustrates the performance improvement. Table \ref{tab:noBalanceVsFullBalanceTableTanh} provides the final accuracy values.}
\label{fig:noBalanceVsFullBalanceTanh}
\end{figure}

\begin{table}[H]
    \centering
    \resizebox{\textwidth}{!}{
        \begin{tabular}{|c|c|c|c|c|c|c|}
        \hline
        Type & \multicolumn{3}{|c|}{No FB at Start} & \multicolumn{3}{c|}{FB at Start} \\ \hline
        & Plain & L1 Reg. & L2 Reg. & Plain & L1 Reg. & L2 Reg.\\ \hline
        2 Layer FCN & 91.69\% & 87.11\% & 91.72\% & \textbf{91.97\%} & \textbf{89\%} & \textbf{91.9\%}
        \\ \hline
        3 Layer FCN & 92.28\% & 82.75\% & 92.03\% & \textbf{92.58\%} & \textbf{90.23\%} & \textbf{92.08\%}
        \\ \hline
        5 Layer FCN & 91.64\% & 87.61\% & 90.99\% & \textbf{92.48\%} & \textbf{91.28\%} & \textbf{91.91\%} \\
        \hline
        \end{tabular}
    }
    \caption{Test accuracy for plain, $L_1$-regularized, and $L_2$-regularized fully connected feedforward networks with $\tanh$ units trained on the MNIST dataset, comparing the effect of full balancing before training to no full balancing before training. Consistently with the results displayed in in Figure \ref{fig:noBalanceVsFullBalanceTanh}, full balancing before training results in faster convergence, as well as higher test accuracy, across all methods. }
    \label{tab:noBalanceVsFullBalanceTableTanh}
\end{table}

In figure \ref{fig:noBalanceVsFullBalanceTanh}, we observe an improvement in both the rate of convergence and the final accuracy when full balancing is applied at the start of training. This is true across the architectures that were testes and across the training methodologies, with and without regularization. Thus balancing at the start of training appears to be useful even when non-BiLU neurons are used. Note that $\tanh$ units behave like BiLU units, and more precisely like linear units, when the weights are very small, i.e., at the beginning of training.

\begin{figure}[H]
\centering
\includegraphics[width=1.\textwidth, keepaspectratio]{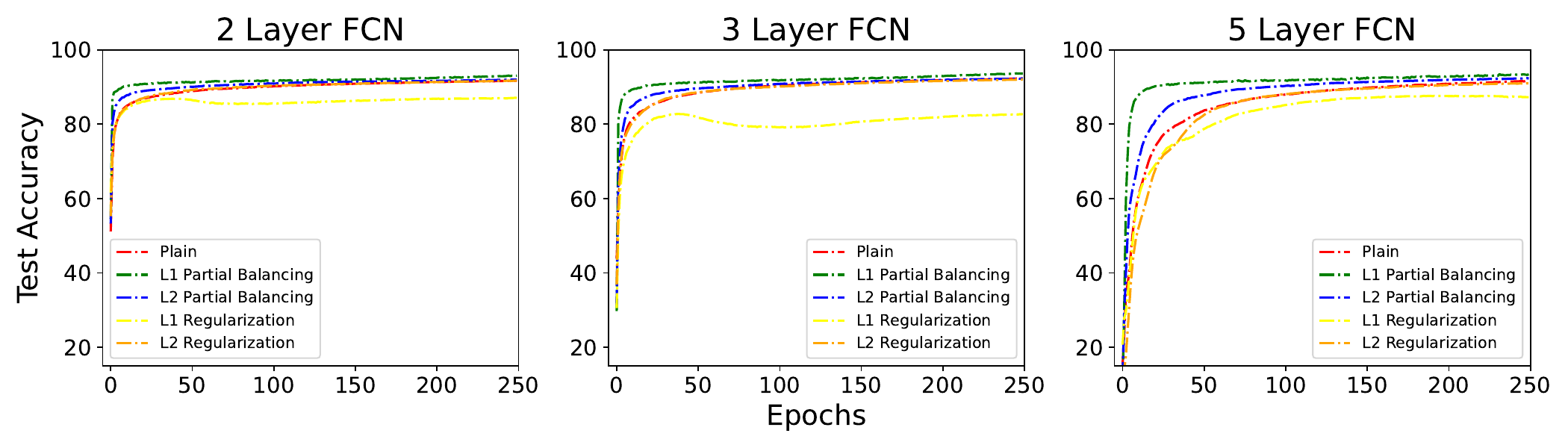}
\caption{Comparison of partial $L_1$ and $L_2$ balancing to $L_1$-regularized and $L_2$-regularized training for different architectures of $\tanh$ units trained on the MNIST dataset.
As the models grow bigger, partial balancing helps the models converge faster and perform better. Table \ref{tab:ComparingMNISTPerformancesTableTanh} reports the final test accuracy values. }
\label{fig:ComparingMNISTPerformancesPlotTanh}
\end{figure}

\begin{table}[H]
    \centering
    \resizebox{0.75\textwidth}{!}{
        \begin{tabular}{|l|c|c|c|c|c|}
        \hline
        Type & Plain & L1 PB & L2 PB & L1 Reg. & L2 Reg.\\
        \hline
        2-FCN & 91.69\% & \textbf{93.07\%} & 92.0\% & 87.11\% & 91.72\% \\
        \hline
        3-FCN & 92.28\% & \textbf{93.66\%} & 92.27\% & 82.75\% & 92.03\% \\
        \hline
        5-FCN & 91.64\% & \textbf{93.44\%} & 92.36\% & 87.61\% & 90.99\% \\
        \hline
        \end{tabular}
    }
    \caption{Comparison of test accuracy across training methods including plain training (no regularization, no balancing), partial balancing (PB) using $L_1$ or $L_2$ costs, and $L_1$  or $L_2$ regularized training using the MNIST dataset for fully connected networks (FCNs) of depth 2, 3, and 5. The networks use $\tanh$ actvation functions 
 As observed in the plots in Figure \ref{fig:ComparingMNISTPerformancesPlotTanh}, $L_1$ partial balancing outperforms the other training methodologies across all model sizes.}
    \label{tab:ComparingMNISTPerformancesTableTanh}
\end{table}

Next we continue with a 
a comparison of partial balancing with other training methodologies on the MNIST dataset, using $\tanh$
activation functions. We observe through Figure \ref{fig:ComparingMNISTPerformancesPlotTanh} and Table \ref{tab:ComparingMNISTPerformancesTableTanh} that partial balancing improves convergence and accuracy. In this case, the best results are obtained with $L_1$ partial balancing. The use of $L_1$ partial balancing significantly improves the rate of convergence and the final test accuracy of the models, irrespective of their size. Again this demonstrates the effectiveness of balancing even in models that use non-BiLU activation functions.
\\

\section{Discussion}

While the theory of synaptic neural balance is a mathematical theory that stands on its own, it is worth considering some of its possible consequences and applications, at the theoretical, algorithmic, biological, and neuromorphic hardware levels.  
%%Some of its applications could be in the theory of deep learning itself. 

\subsection{Theory}
Theories of deep learning in networks of McCulloch and Pitt neurons must often proceed by fixing the kinds of activation functions and neurons that are being used. At one extreme end of the spectrum, one finds linear networks for which a rich theory is available
\cite{baldi88,baldi2021deep}. At the other most non-linear end of the spectrum, one finds networks of unrestricted Boolean functions which are also amenable to fairly general analyses \cite{baldijmlr12}. In between, one typically considers networks of linear threshold neurons
\cite{baldi2019capacity}, or sigmoidal neurons (logistic or tanh activation functions), or ReLU neurons \cite{petersen2018optimal}. The results shown here suggest that results obtained for networks of linear or ReLU neurons, may be exttendable to networks of BiLU neurons, especially when the homogeneity property plays an essential role, and perhaps also other neurons such as RePU neurons. 
In short, a line of investigation suggested by this work to study all the basic questions about deep learning (e.g. capacity, generalization, universal approximation properties) in networks of BiLU neurons of increasing architectural complexity. It is easy to show, for instance, that BiLU networks have universal approximation properties (see Appendix).

Our results show that for a given architecture with weights $W$, there is an entire equivalence class of weights with the same overall performance, associated with scaling operations and the underlying linear manifold. Global balancing can be viewed as a systematic way of selecting a unique canonical representative within the class, associated with the corresponding balanced network.
Among other things, the existence of such equivalence classes implies that the information in the training data does not need to be able to specify the individual weights, but may instead specify the equivalence class of the weights. This is tied to the formal notion 
of capacity \cite{baldi2019capacity} and explain in part why large networks can be trained with less than $\vert W \vert$ examples while not overfitting \cite{baldi2024deep}.  It is worth noting that the equivalence classes corresponding to weights that provide the same input-output function may be even larger than what is given by the linear manifold of scalings, as they could contain other operations besides scaling, such as permuting the neurons of any given layer in the fully connected case. However, in the case of a feedforward network of BiLUs where the units are numbered, and connections run only from lower-numbered units to higher-numbered units, then each unit has its unique connectivity pattern and the equivalence classes may be restricted to scaling operations.  

Another theoretical application is the study of learning in linear networks with $L_p$ regularization. For instance, it is well known that a feedforward, fully connected, linear autoencoder with bottleneck layers trained to minimize the sum of square reconstruction error $E$ has a unique global optimum, up to trivial transformations,  corresponding to Principal Component Analysis (PCA) in the bottleneck layers. Furthermore, that problem has no spurious local minima and all other critical points of the error function are saddle points associated with projections onto linear spaces spanned by the non-principal components. What happens when an $L_p$ regularizer $R$ is added to the reconstruction error, so that the overall error is given by ${\ E} =E + \beta R$. For $\beta$ very large, $R$ will dominate and the optimal solution is to have all the weights equal to 0. However, when $\beta$ is small, the optimal solution is provided by the theory presented here. The error is dominated by $E$ so the optimum is associated with the PCA solution, as described above, which can then be refined by balancing to further reduce $R$ and reach the global optimum.  

\subsection{Algorithms}

The theory and the simulations in Section 
\ref{sec:experiments} show that synaptic balance has a number of different applications. It can be used:
(1) To initialize the weights of a network at the start of training to better condition the learning;
(2) To check that a network trained to minimize $E+R$ has been properly trained by checking that each neuron is balanced with respect to $R$; (3) To balance a non-properly trained network at the end of training; 
(4) To better regularize a network by alternating SGD steps applied to $E$, or even $E+R$, with partial or full balancing steps. This may improve the convergence or final performance of learning;  (5) this of course can be applied to all situations where regularization is beneficial, including the case of smaller training sets; and (6) To tweak a network trained with ah $L_p$ regularization, towards a state where it is $L_q$ regularized (with $q \not = p$) without retraining it. 
Thus, in short, balancing is a little bit like dropout: it ought to be added to the arsenal of tools available to improve neural network training.

\subsection{Biology}
The balancing operations are {\it local} \cite{baldi2016local}, in the sense that they involve only the pre-and post-synaptic weights of a neuron. Thus, unlike backpropagation, balancing is plausible in a physical neural system, as opposed to a digitally simulated neural system. Thus synaptic balancing could be of interest in neuroscience or in neuromorphic engineering, for instance from the standpoint of learning or memory maintenance. While there is extensive literature in neuroscience on the balance between excitation and inhibition in biological networks  \cite{van1996chaos,tatti2017neurophysiology}, there is little evidence in favor or against neuronal synaptic balance in the sense described here. However, there is some evidence for the existence of homeostatic processes that scale the input synaptic weights to regulate the activity of neurons \cite{turrigiano2012homeostatic,chistiakova2015homeostatic,turrigiano2017dialectic}. In addition, there is also evidence that biological neurons can scale their intrinsic excitability (spike threshold) to regulate their activity
\cite{huang2016adaptive,fontaine2014spike}. It is at least conceivable that these two scaling mechanisms could be at play and work together in some situations
(see also \cite{stock2022synaptic}). In any case, current technology is quite far from allowing one to measure the strength of all the incoming and outgoing synapses of a biological neuron in an animal brain. Thus, it is difficult to draw any definitive conclusions on the existence of some kind of homeostasis between the incoming and outgoing synapses of biological neurons. Exploratory experiments could potentially be carried either in simpler organisms with a small number of well-defined neurons, such as {\it C. elegans}, or in cultured neurons. In addition, simulations could be carried out in more detailed compartmental neural network models,
possibly ones where inhibitory and excitatory neurons are segregated in uneven proportions--the overall ratio of excitatory to inhibitory neurons in the mammalian cortex is roughly 80\% to 20\% 
\cite{rubenstein2003model}, with biologically-observed connectivity patterns.
Finally, as a most speculative and for now untestable hypothesis, one may conjecture that biological neurons may undergo balancing phases, and that perhaps those phases occur during the night, with a close association between synaptic balancing and dreaming. 

\subsection{Neuromorphic Hardware}

In physical neural networks (e.g. biological neural networks, neuromorphic chips), as opposed to digitally simulated neural networks, the algorithms for adjusting synaptic weights must be local both in space and time. Yet these networks must exhibit good global properties. Thus, algorithms that are local and lead to global order, such as synaptic balancing, are of particular interest when considering physical, non-simulated, neural systems.
For example, while a deep neuron embedded in a physical network may have no way of monitoring the global training error, conceivably it could sense and monitor its degree of balance, since that is an entirely local property. The degree of balance could be used as a local proxy to monitor global learning progress.  Furthermore, the physical properties of the underlying hardware could constrain the synaptic weights or the learning algorithms in ways that could require or benefit from some form of synaptic balance. 
%%whether synaptic balance could have more practical applications either in learning or memory maintenance; and whether synaptic balance may play a role in biological or neuromorphic circuits.

Optimization strategies for training neuromorphic spiking neural networks with low energy consumption are analyzed in \cite{sorbaro2020optimizing} (see also \cite{rueckauer2017conversion}), including how the homogeneous properties of ReLU neurons
can influence the number of spikes generated at each layer and the average energy consumption at each layer. 
Thus there is a direct connection between synaptic balance and some of the neuromorphic literature using scaling methods to control spiking rates and energy consumption. In particular, balancing may help minimize energy consumption.
Similar considerations apply to memristor-based neuromorphic hardware 
\cite{ivanov2022neuromorphic, 
ji2016neutrams}.

\section{Conclusion}
The theory of synaptic neural balance explains some basic findings regarding $L_2$ balance in feedforward networks of ReLU neurons and extends them in several directions. The first direction is the extension to BiLU and other activation functions (BiPU). The second direction is the extension to more general regularizers, including all $L_p$ ($p>0$) regularizers. The third direction is the extension to non-layered architectures, recurrent architectures, convolutional architectures, as well as architectures with mixed activation functions. The theory is based on two local neuronal operations: scaling which is commutative, and balancing which is not commutative. 
Finally, and most importantly, given any initial set of weights, when local balancing operations are applied in a stochastic or deterministic manner, global order always emerges through the convergence of the balancing algorithm to the same unique set of balanced weights. The reason for this convergence is the existence of an underlying convex optimization problem where the relevant variables are constrained to a linear, only architecture-dependent, manifold.  
Balancing can be applied, fully or partially, before, during, or at after training and can help improve the convergence or final performance. Scaling and balancing operations are local and thus may have applications in physical, non-digitally simulated, neural networks where the emergence of global order from local operations may lead to better operating characteristics and lower energy consumption. 

%%\section*{Acknowledgment}
\null\par
\noindent
{\bf Contributions:} The theory was developed by PB. The simulations to corroborate the theory were carried out by A. R. 
The simulations to improve training were carried out by I.D. and A. A. The manuscript was written by P.B with input from all the other authors. All authors reviewed the final version.

%%\clearpage

\section*{Appendix A: Universal Approximation Properties of BiLU Neurons}
Here we show that any continuous real-valued function defined over a compact set of the Euclidean space can be approximated to any degree of precision by a network of BiLU neurons with a single hidden layer. 
As in the case of the similar proof given in \cite{baldi2021deep} using linear threshold gates in the hidden layer, it is enough to prove the theorem for a continuous function $f$: $ {0,1} \to \R$.

\begin{theorem} {(Universal Approximation Properties of BiLU Neurons)}
Let $f$ be any continuous function from $[0,1]$ to $\R$ and $\epsilon >0$. Let $g_\lambda$ be the ReLU activation function with slope $\lambda \in \R$s. Then there exists a feedforward network with a single hidden layer of neurons with ReLU activations of the form $g_\lambda$ and a single output linear neuron, i.e., with BiLU activation equal to the identity function, capable of approximating $f$ everywhere within $\epsilon$ (sup norm).
\end{theorem}

\begin{proof}
To be clear, $g_\lambda(x)=0$ for $x<0$ and $g_\lambda(x)=\lambda x$ for $0 \leq x$.
Since $f$ is continuous over a compact set, it is uniformly continuous. Thus there exists $\alpha>0$ such that for any $x_1$ and $x_2$ in the $[0,1]$ interval:

\be
\vert x_2-x_1 \vert < \alpha  \implies \vert f(x_2) -f(x_1) \vert < \epsilon\
\label{universal101}
\ee
Let $N$ be an integer such that $1<N\alpha$, and let us slice the interval $[0,1]$ into $N$ consecutive slices of 
width $h=1/N$, so that within each slice the function $f$ cannot jump by more than $\epsilon$. Let us connect the input unit to all the hidden units with a weight equal to 1. Let us have $N$ hidden units numbered $1, \ldots, N$ with biases equal to $0, 1/N,2/N,....,N_1/N$ respectively and activation function of the form $g_{\lambda_k}$. It is essential that different units be allowed to have different slopes $\lambda_k$. The input unit is connected to all the hidden units and all the weights on these connections are equal to 1. Thus when $x$ is in the $k$-th slice,  $(k-1)/N \leq x  < k/N$, all the units from $k+1$ to $N$ have an output equal to $0$, and all the units from 1 to $k$ have an output determined by the corresponding slopes. All the hidden units are connected to the output unit with weights $\beta_1, \ldots, \beta_N$, and $\beta_0$ is the bias of the output unit.
We want the output unit to be linear. In order for the $\epsilon$ approximation to be satisfied, it is sufficient if in the 
$(k-1)/N \leq x < k/N$ interval, the output is equal to the line 
joining the point $f((k-1)/N)$ to the point $f(k/N)$. In other words, if $x\in [(k-1)/N, k/N)$, then we want the output of the network to be:

\be
\beta_0+ \sum_{i=1}^k \beta_i \lambda_i (x-(i-1)h)
=
f(\frac{k-1}{N}) + \frac{ f(\frac{k}{N})-f (\frac{k-1}{N}) }{h} (x-(k-1)h)
\label{eq:lines}
\ee
By equating the y-intercept and slope of the lines on the left-hand side and the righ- hand side of Equation \ref{eq:lines}, we can solve for the weights $\beta $'s and the slopes $\lambda$'s.
\end{proof}
As in the case of the similar proof using linear threshold functions in the hidden layer (see \cite{baldi2021deep},) this proof can easily be adapted to continuous functions defined over a compact set of $\R^n$, even with a finite number of finite discontinuities, and into 
$\R^m$.

\section*{Appendix B: Analytical Solution for the Unique Global Balanced State}

Here we directly prove the convergence of stochastic balancing to a unique final balanced state, and derive the equations for the balanced state, in the special case of tied layer balancing (as opposed to single neuron balancing). The Proof and the resulting equations are also valid for stochastic balancing (one neuron at a time) in a layered architecture comprising a single neuron per layer. 
Let us call tied layer scaling the operation by which all the incoming weights to a given layer of BiLU neurons are multiplied by $\lambda>0$ and all the outgoing weights of the layer are multiplied by $1/\lambda$, again leaving the training error unchanged. Let us call layer balancing the particular scaling operation corresponding to the value of $\lambda$ that minimizes 
the contribution of the layer to the $L_2$ (or any other $L_p$) regularizer vaue. This optimal value of $\lambda^*$ results in layer-wise balance equations: the sum of the squares of all the incoming weights of the layer must be equal to the sum of the squares of all the outgoing weights of the layer in the $L_2$ case, and similarly in all $L^P$ cases. 

\begin{theorem}  
Assume that tied layer balancing is applied iteratively and stochastically to the layers of a layered feedforward network of BiLU neurons.  As long as all the layers are visited periodically, this procedure will always converge to the same unique set of weights, which will satisfy the layer-balance equations at all layers, irrespective of the details of the schedule. Furthermore,
the balance state can be solved analytically. 
\end{theorem}

\begin{proof}
Every time a layer balancing operation is applied, the training error remains the same, and the $L_2$ (or any other $L_p$) regularization error decreases or stays the same. Since the regularization error is always positive, it must converge to a certain value. Using the same arguments as in the proof of Theorem 
\ref{thm:uniqueness},  the weights must also converge to a stable configuration, and since the configuration is stable all its layers must satisfy the layer-wise balance equation. The key remaining question is why is this configuration unique and can we solve it analytically? Let $A_1,A_2, \ldots A_N$ denote the matrices of connections between the layers of the network.  Let $\Lambda_1,\Lambda_2, \ldots,\Lambda_{N-1}$ be $N-1$ strictly positive multipliers, representing the limits of the products of the corresponding $\lambda_i^*$ associated with each balancing step at layer $i$, as in the proof of Theorem \ref{thm:uniqueness}. In this notation, layer 0 is the input layer and layer $N$ is the output layer (with $\Lambda_0=1$ and $\Lambda_N=1$).

After converging, each matrix $A_i$ becomes the matrix
$ \Lambda_i/\Lambda_{i-1} A_i =M_iA_i $ for $i=1 \ldots N$, with $M_i=\Lambda_i/\Lambda_{i-1}$.
The multipliers $M_i$ must minimize the regularizer while satisfying $M_1 \ldots M_N=1$ to ensure that the training error remains unchanged.  
In other words, to find the values of the  $M_i$'s we must minimize the Lagrangian:

\be
{\mathcal {L}} (M_1, \ldots,M_N)=
\sum_{i=1}^N  \vert \vert M_iA_i \vert \vert^2 + \mu (1-\prod_{i=1}^N M_i)
\ee
written for the $L^2$ case in terms of the Frobenius norm, but the analysis is similar in the general $L_p$ case. From this, we get the critical equations:

\be
\frac{\partial {\mathcal{ L}}}
{\partial M_i}=2M_i \vert \vert A_i \vert \vert^2 -\mu M_1 \ldots M_{i-1}M_{i+1} \ldots M_N=0
\quad {\rm for} \; i=1, \ldots,N  \quad {\rm and} \quad  \prod_{i=1}^N  M_i=1
\ee
As a resut, for every $i$:

\be
2M_i \vert \vert A_i \vert \vert^2 -\frac{\mu }{M_i} =0
\quad {\rm or} \quad \mu= 2 M_i^2 \vert \vert A_i \vert \vert^2
\ee
Thus each $M_i>0$ can be expressed in a unique way as a function of the Lagrangian multiplier $\mu$ as:
$M_i = (\mu/2 \vert \vert A_i \vert \vert^2)^{1/2}$.
By writing again that the product of the $M_i$is equal to 1, we finally get:

\be
\mu^N=2^N \prod_{i=1}^N \vert \vert A_i \vert \vert^2 \quad {\rm or}
\quad \mu=2 \prod_{i=1}^N \vert \vert A_i \vert \vert^{2/N}
\ee
Thus we can solve for $M_i$:

\be
M_i=\frac{\mu}{2 \vert \vert A_i \vert\vert^2}=
\frac{\prod_{i=1}^N \vert \vert A_i \vert \vert^{2/N}}{\vert \vert A_i \vert \vert^2}
\quad \quad {\rm for} \;\; i=1, \ldots, N
\label{eq:lagrange101}
\ee
Thus, in short, we obtain a unique closed-form expression for each  $M_i$. From there, we infer the unique and final state of the weights, where 
$ A^*_i=  M_iA_i$.
Note that each $M_i$ depends on all the other $M_j$'s, again showcasing how the local balancing algorithm leads to a unique global solution. 
\end{proof}

\clearpage
\bibliography{baldi,nn} 
\bibliographystyle{plain}

\end{document}